\documentclass[10pt,twocolumn,letterpaper]{article}

\usepackage{iccv}              

\definecolor{iccvblue}{rgb}{0.21,0.49,0.74}
\usepackage[pagebackref,breaklinks,colorlinks,allcolors=iccvblue]{hyperref}

\usepackage{adjustbox}
\usepackage{xspace}
\usepackage{comment}
\usepackage{pifont}
\usepackage{makecell}
\usepackage{colortbl}
\usepackage{xcolor}
\usepackage{titletoc} 
\usepackage[most]{tcolorbox}
\usepackage{wrapfig}
\usepackage{longtable}
\usepackage{fontawesome5}
\usepackage{amsmath}
\usepackage{comment}

\newcommand{\datasetName}{\textsc{MultiVerse}\xspace}

\newcommand{\verification}{\raisebox{-0.5ex}{\includegraphics[height=1em]{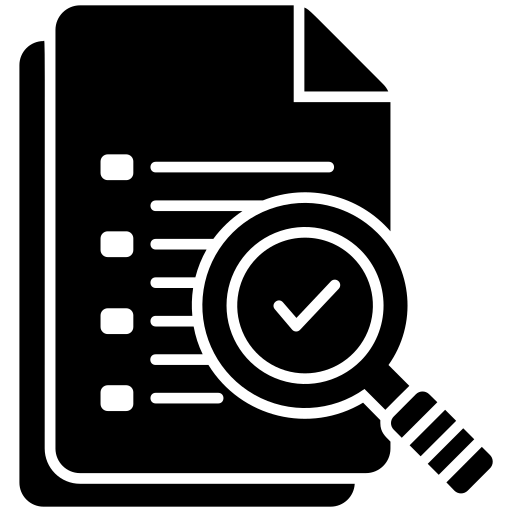}}}
\newcommand{\analysis}{\raisebox{-0.5ex}{\includegraphics[height=1em]{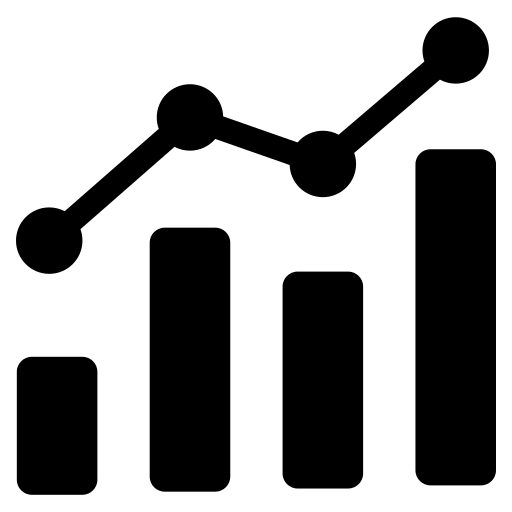}}}
\newcommand{\exploration}{\raisebox{-0.5ex}{\includegraphics[height=1em]{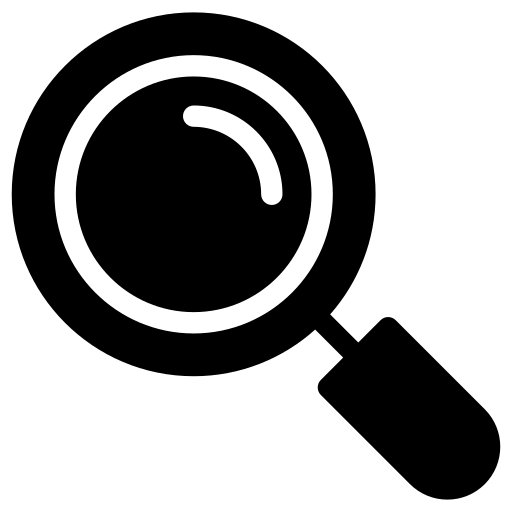}}}
\newcommand{\optimization}{\raisebox{-0.5ex}{\includegraphics[height=1em]{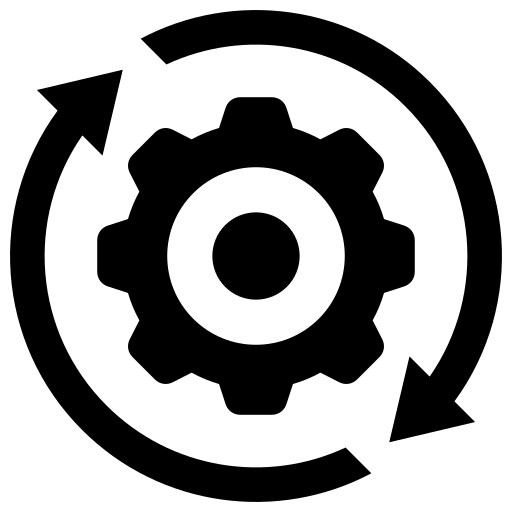}}}
\newcommand{\calculation}{\raisebox{-0.5ex}{\includegraphics[height=1em]{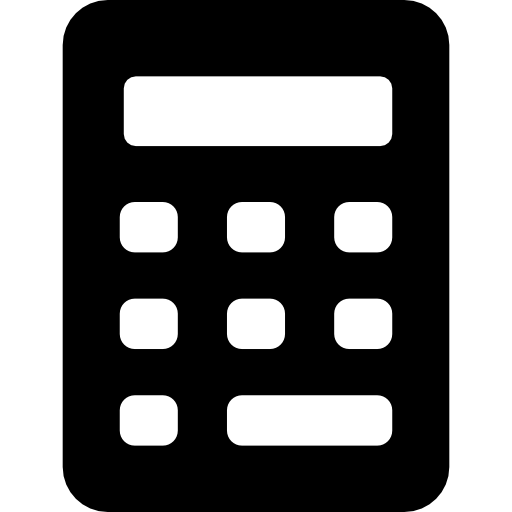}}}
\newcommand{\understanding}{\raisebox{-0.5ex}{\includegraphics[height=1em]{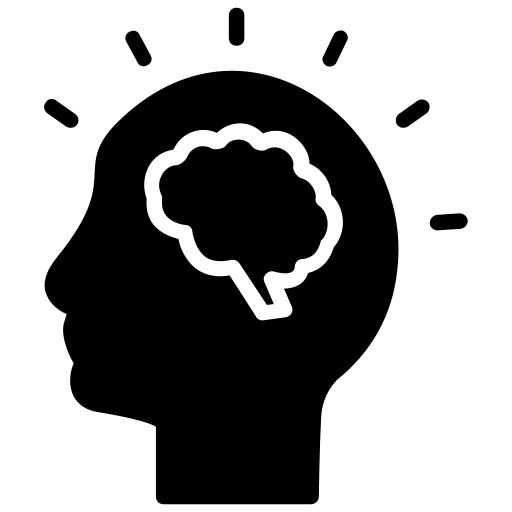}}}
\newcommand{\research}{\raisebox{-0.5ex}{\includegraphics[height=1em]{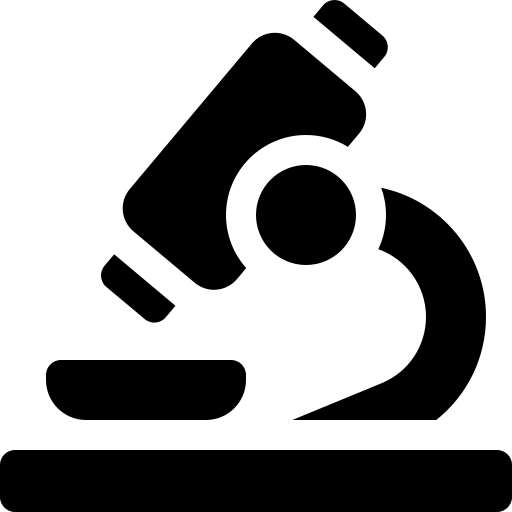}}}
\newcommand{\creation}{\raisebox{-0.5ex}{\includegraphics[height=1em]{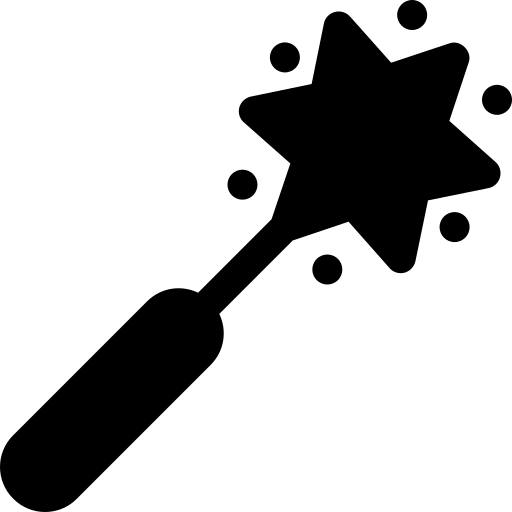}}}

\newcommand{\github}[1]{%
   \href{#1}{\faGithub}%
}

\newtcolorbox{prompt}[2][]{
    colback=white,
    colframe=gray!45,
    fonttitle=\bfseries,
    coltitle=black,
    title=#2,
    #1,breakable
}

\def\paperID{11756} 
\def\confName{ICCV}
\def\confYear{2025}

\title{\raisebox{-0.2cm}{\includegraphics[width=0.8cm]{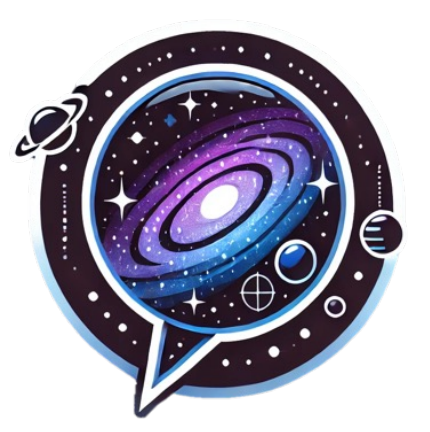}}\datasetName: A Multi-Turn Conversation Benchmark for Evaluating Large Vision and Language Models}

\author{Young-Jun Lee \textsuperscript{\rm $\heartsuit$} ~Byung-Kwan Lee \textsuperscript{\rm $\heartsuit$} ~Jianshu Zhang \textsuperscript{\rm $\nabla$} \\
Yechan Hwang \textsuperscript{\rm $\heartsuit$} ~
Byungsoo Ko \textsuperscript{\rm $\heartsuit$} ~
Han-Gyu Kim \textsuperscript{\rm $\clubsuit$} ~
Dongyu Yao \textsuperscript{\rm $\diamondsuit$} ~
Xuankun Rong \textsuperscript{\rm $\nabla$} ~ \\
Eojin Joo \textsuperscript{\rm $\heartsuit$} ~
Seung-Ho Han \textsuperscript{\rm $\heartsuit$} ~
Bowon Ko \textsuperscript{\rm $\heartsuit$} ~
Ho-Jin Choi \textsuperscript{\rm $\heartsuit$}
\\\\
\textsuperscript{\rm $\heartsuit$} KAIST ~
\textsuperscript{\rm $\nabla$} WHU ~
\textsuperscript{\rm $\clubsuit$} NAVER ~
\textsuperscript{\rm $\diamondsuit$} CMU ~
\\
}

\begin{document}

\twocolumn[{
\maketitle
\vspace{-30pt}
\begin{center}
    \centering
    \captionsetup{type=figure}
    \vspace{5mm}
    \includegraphics[width=1.0\textwidth]{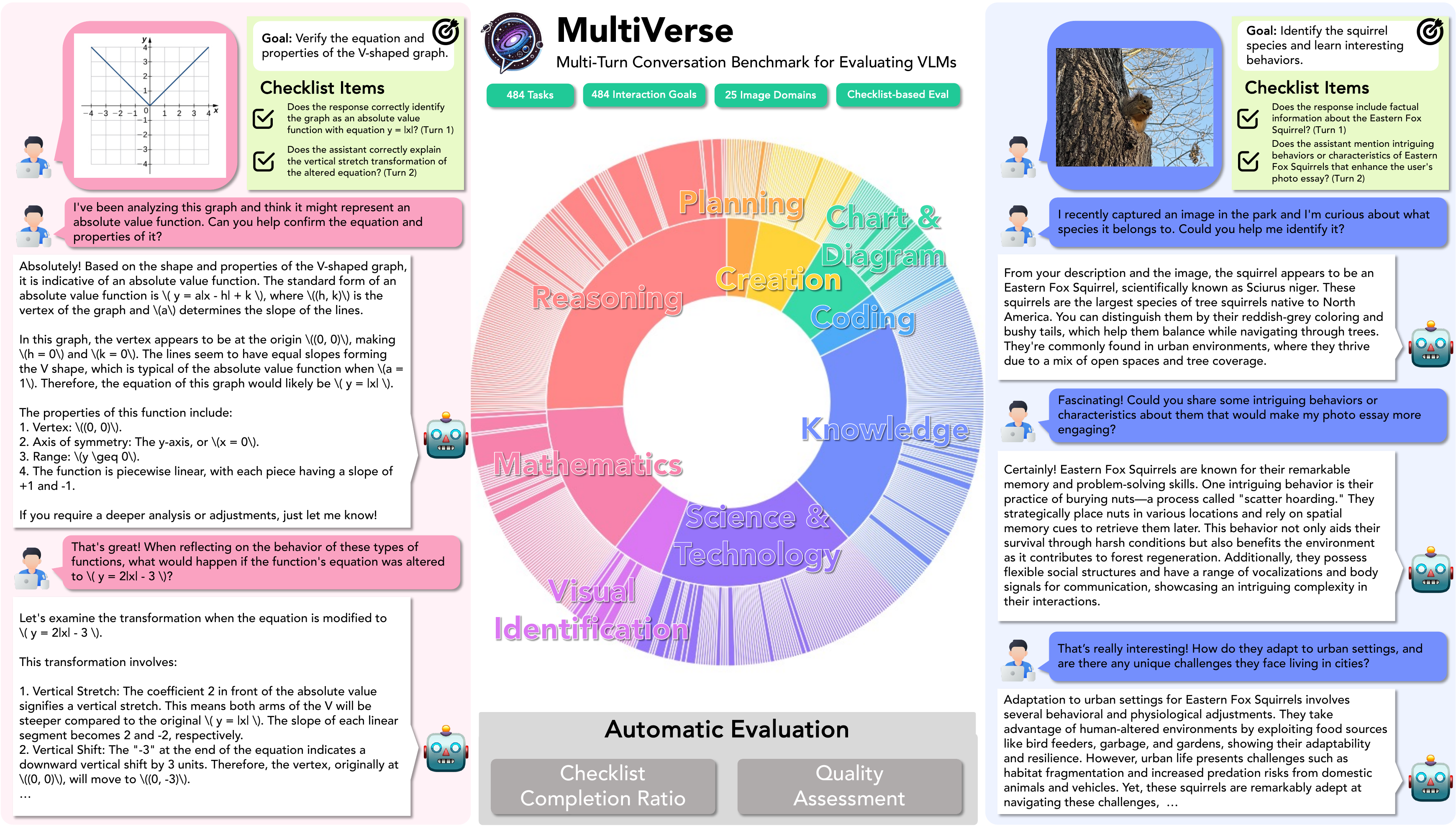}
  \caption{An overview of \datasetName, a multi-turn conversation benchmark designed to evaluate VLMs across eight main tasks (\eg Reasoning, Mathematics, Knowledge) comprising 484 tasks, nine main interaction goals (\eg Verification, Exploration) with 484 distinct interaction goals, and 25 diverse image domains (\eg Charts and Graphs, Nature). \datasetName provides instance-specific checklist evaluation items for each turn and is the first multi-turn conversation benchmark encompassing diverse and challenging tasks.}
  \label{main_fig:teaser}
  \vspace{3mm}
\end{center}
}]

\begin{abstract} 
Vision-and-Language Models (VLMs) have shown impressive capabilities on single-turn benchmarks, yet real-world applications often demand more intricate multi-turn dialogues. Existing multi-turn datasets (\eg MMDU, ConvBench) only partially capture the breadth and depth of conversational scenarios encountered by users. In this work, we introduce \datasetName, a novel multi-turn conversation benchmark featuring 647 dialogues—each averaging four turns—derived from a diverse set of 12 popular VLM evaluation benchmarks. With 484 tasks and 484 interaction goals, \datasetName covers a wide range of topics, from factual knowledge and perception to advanced reasoning tasks such as mathematics and coding. To facilitate robust assessment, we propose a checklist-based evaluation method that leverages GPT-4o as the automated evaluator, measuring performance across 37 key aspects, including perceptual accuracy, linguistic clarity, and factual correctness. We evaluate 18 VLMs on \datasetName, revealing that even the strongest models (e.g., GPT-4o) achieve only a 50\% success rate in complex multi-turn conversations, highlighting the dataset's challenging nature. Notably, we find that providing full dialogue context significantly enhances performance for smaller or weaker models, emphasizing the importance of in-context learning. We believe \datasetName is a landscape of evaluating multi-turn interaction abilities for VLMs. We make our source code and dataset
publicly available.~\footnote{\url{https://passing2961.github.io/multiverse-project-page/}}
\end{abstract}

\section{Introduction} \label{sec:intro}

Vision-and-Language Models (VLMs)~\cite{gpt4o,anthropic2024claude,Qwen2.5-VL} have recently demonstrated remarkable performance on various evaluation benchmarks~\cite{fu2023mme,lu2023mathvista,yu2023mm,yu2024mm,yue2024mmmu,yue2024mmmupro,chen2024far,li2024naturalbench}. These benchmarks primarily assess the perception and reasoning capabilities of VLMs through simple binary, multiple-choice questions, or short free-form responses in single-turn interactions. However, real-world usage often involves users engaging in continuous, multi-turn conversations with models to solve problems iteratively. While multi-turn evaluation benchmarks have introduced for evaluating LLMs~\cite{wang2023mint,zheng2023judging,bai2024mt,kwan2024mt,he2024multi,sirdeshmukh2025multichallenge,lin2024wildbench}, comparatively few benchmarks (\ie MMDU~\cite{liu2024mmdu}, ConvBench~\cite{liu2024convbench}) address VLMs’ abilities in multi-turn conversation. This gap raises the critical question: \textbf{Can VLMs that excel in single-turn benchmarks also meet user needs in more interactive, multi-turn scenarios?}

\begin{figure}[!t]
    \centering
    \includegraphics[width=\linewidth]{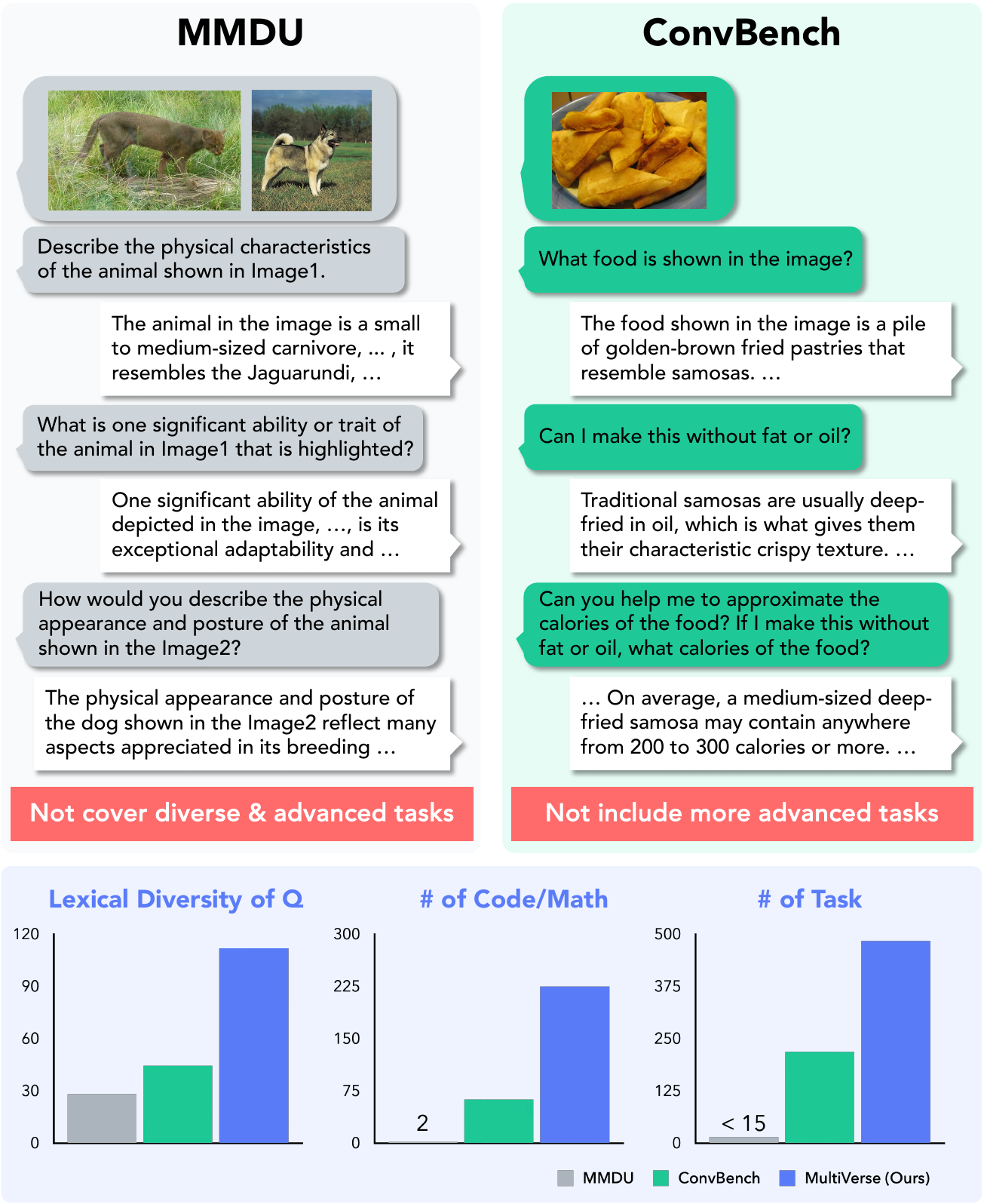}
    \vspace{-1em}
    \caption{Examples of multi-turn conversations from MMDU~\cite{liu2024mmdu} and ConvBench~\cite{liu2024convbench}, along with their limitations. The graph below compares MMDU, ConvBench, and \datasetName in terms of query lexical diversity, the number of code/math tasks in responses, and the total number of tasks.} 
    \label{main_fig:problem}
    \vspace{-1em}
\end{figure}

Despite recent efforts to address this issue, existing benchmarks still exhibit significant shortcomings and do not adequately cover VLMs’ capabilities in multi-turn interactions, particularly across diverse tasks (breadth) and reasoning requirements (depth). As shown in Figure~\ref{main_fig:problem}, primarily features knowledge-oriented images (\eg landscapes, animals, art) derived from WIT~\cite{srinivasan2021wit}, placing substantial emphasis on knowledge-specific reasoning tasks (\eg ``\textit{Describe the object in the image}''). Hence, MMDU offers a relatively narrow set of fewer than 15 tasks and a limited range of question styles with low lexical diversity. In contrast, ConvBench encompasses three main task categories (\ie perception, reasoning, creation) and covers 219 sub-tasks derived from VisITBench~\cite{bitton2023visit}. This design significantly broadens its task coverage. However, it lacks more advanced tasks, such as \textit{equation verification} in Figure~\ref{main_fig:teaser}. Furthermore, code- and mathematics-related responses remain sparse, and users’ queries are comparatively simple in their linguistic structure. These shortcomings underscore the need for a more comprehensive and holistic multi-turn conversation benchmark over more diverse and challenging tasks (breadth $\uparrow$ \& depth $\uparrow$).

\begin{figure*}[!t]
    \centering
    \includegraphics[width=\linewidth]{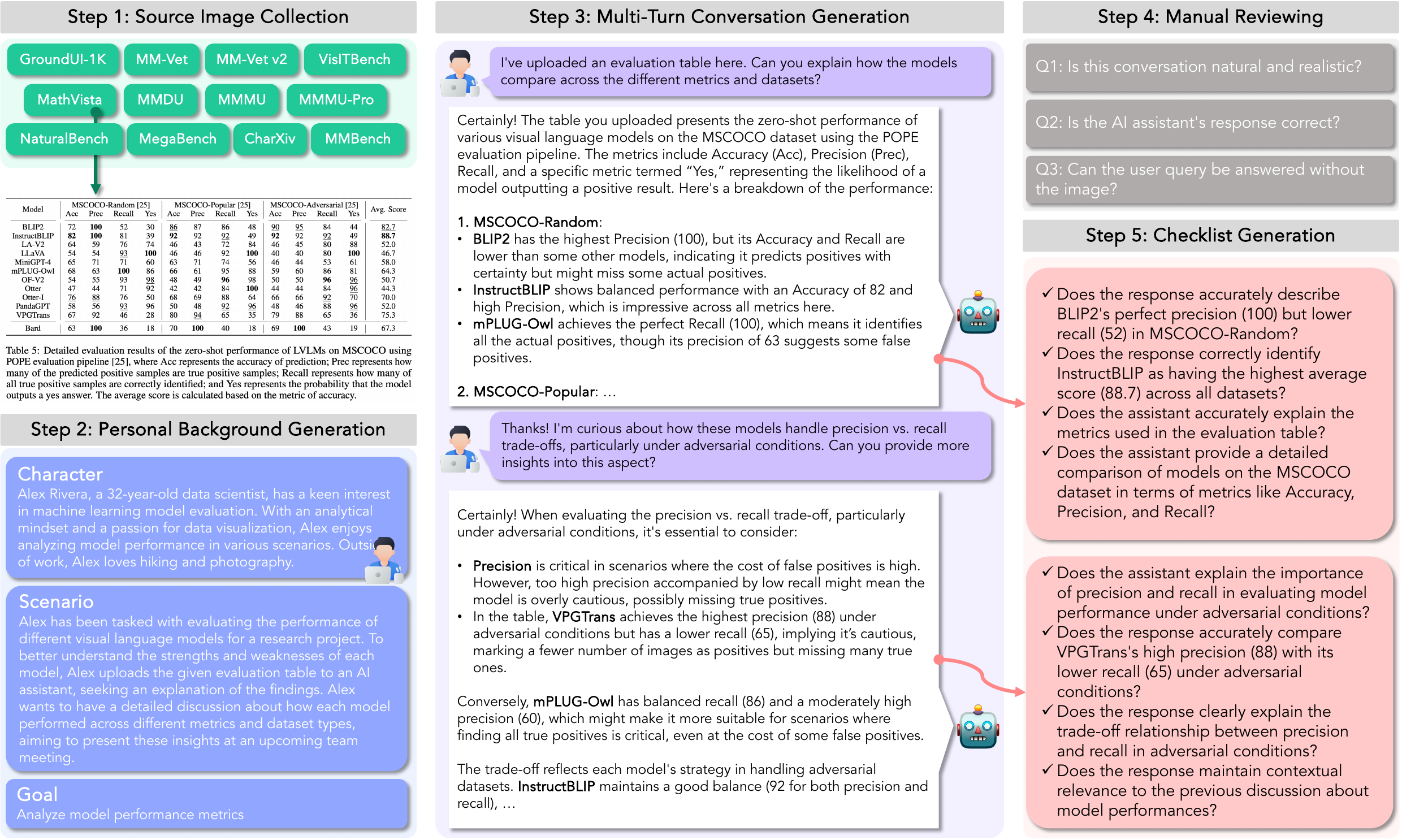}
    \caption{An overview of \datasetName construction pipeline, consisting of five steps: (1) Source Image Collection, gathering images from various benchmarks; (2) Personal Background Generation, creating user personas with specific goals; (3) Multi-Turn Conversation Generation, where an AI assistant analyzes model performance based on evaluation tables; (4) Manual Reviewing, assessing response correctness and relevance; and (5) Checklist Generation, ensuring quality and coherence in AI responses.}
    \label{main_fig:pipeline}
    \vspace{-1em}
\end{figure*}

In this paper, we introduce \datasetName, a novel multi-turn conversation benchmark designed to evaluate VLMs on more diverse and advanced user tasks. \datasetName comprises 647 conversations, each averaging around four turns, spanning 484 tasks (\eg reasoning, mathematics) and 484 interaction goals (\eg verification, analysis). We first collect source images from 12 widely used VLM evaluation benchmarks (\eg MegaBench~\cite{chen2024mega}, CharXiv~\cite{wang2024charxiv}, MMMU~\cite{yue2024mmmu}), covering a broad range of domains such as nature, science, and mathematics. After selecting high-quality seed images, we adopt a personal background-to-conversation approach~\cite{kim2022soda,jang2023conversation,lee2024stark}, which ensures high lexical diversity-based multi-turn dialogues. Finally, we meticulously remove any unsuitable samples through manual review, focusing on natural conversation flow, factual correctness, and blindness criteria~\cite{fu2024blink,li2024naturalbench}. Therefore, as shown in Figure~\ref{main_fig:problem}, \datasetName encompasses a diverse range of tasks, enabling the handling of more advanced reasoning capabilities (\eg mathematics, coding) to address increasingly complex user queries with higher lexical diversity. For the automatic evaluation of \datasetName, we adopt a checklist-based evaluation metric inspired by prior works~\cite{lee2024checkeval,lin2024wildbench}, which utilize checklist items to assess open-ended generation tasks. Specifically, we employ GPT-4o as the evaluator VLM to measure response quality. Each checklist consists of multiple binary questions covering 37 key aspects, including perceptual understanding and factual correctness, as illustrated in Figure~\ref{main_fig:teaser}.

We evaluate 18 VLMs, including both open-source and proprietary models, on \datasetName. Our findings reveal that most VLMs struggle with multi-turn interactions, with even the strongest model, GPT-4o, achieving less than 50\% performance. This highlights a fundamental challenge in sustained dialogue reasoning.
Interestingly, providing ground-truth dialogue history significantly enhances reasoning capabilities, even for smaller or weaker VLMs—demonstrating a clear in-context learning effect that enables models to adapt and refine their responses over time. Additionally, scaling up model size consistently improves multi-turn interactivity, reinforcing the importance of model capacity in complex conversational tasks.
Finally, we verify that our evaluation metrics remain robust against verbosity bias, ensuring reliable and fair assessment across models. These insights shed light on the limitations of current VLMs and underscore the necessity of improving contextual understanding in long-form interactions.

\begin{figure*}[!t]
    \centering
    \includegraphics[width=\linewidth]{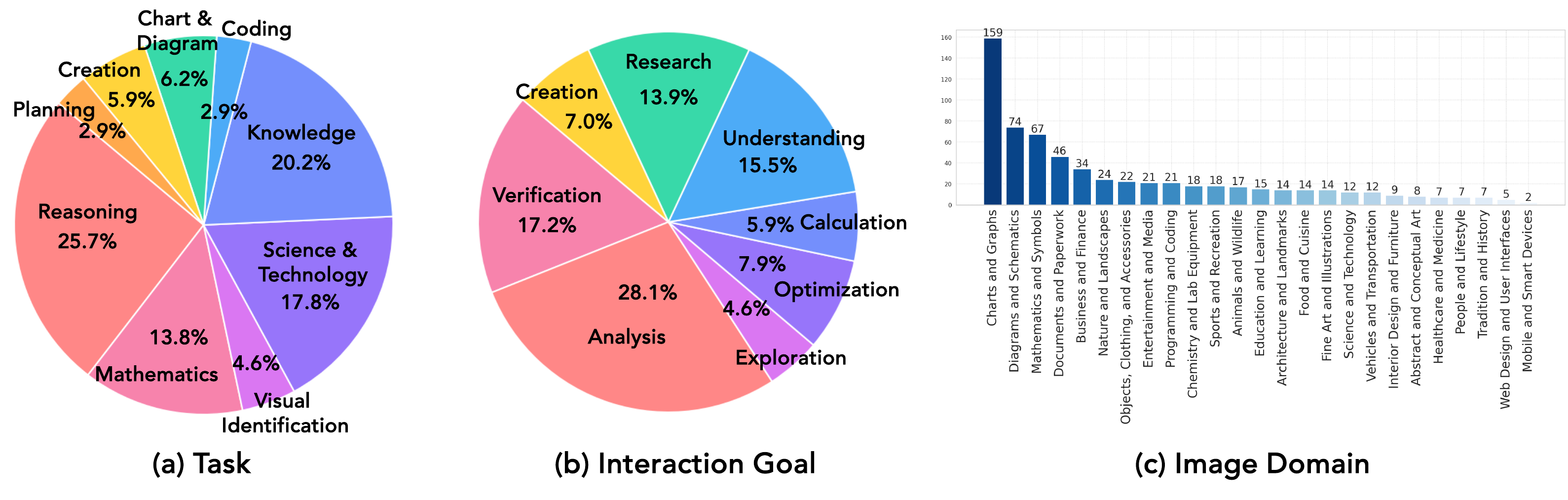}
    \caption{Detailed distribution of (a) Task, (b) Interaction Goal, and (c) Image Domain in \datasetName.}
    \vspace{-1em}
    \label{main_fig:detailed_distribution}
\end{figure*}

\section{\datasetName} \label{main_sec:dataset}

In this section, we will describe how we construct \datasetName, with an overview shown in Figure~\ref{main_fig:pipeline}. 

\subsection{Dataset Construction} \label{main_sec:dataset_construction}

\paragraph{Step 1: Source Image Collection.} 
We began by collecting source images spanning diverse topics (\eg science, mathematics, and natural scenes) from 12 existing evaluation benchmarks: MMDU~\cite{liu2024mmdu}, GroundUI-1K~\cite{zheng2024agentstudio}, MMMU~\cite{yue2024mmmu}, MMMU-Pro~\cite{yue2024mmmupro}, NaturalBench~\cite{li2024naturalbench}, VisIT-Bench~\cite{bitton2023visit}, MathVista~\cite{lu2023mathvista}, MM-Vet~\cite{yu2023mm}, MM-Vet v2~\cite{yu2024mm}, CharXiv~\cite{wang2024charxiv}, MMBench (Eng)~\cite{liu2023mmbench}, and MegaBench~\cite{chen2024mega}. In total, 49,700 images were obtained from the test sets of these benchmarks. We then applied a rigorous three-step data refinement process to ensure high-quality selections: \textbf{(1) Deduplication}: We used pHash to detect and remove duplicate images, resulting in 13,998 images (29.1\%) being discarded; \textbf{(2) Image Quality Scoring}: We employed GPT-4o-2024-11-20 to rate each image on a 1--5 scale in terms of clarity, resolution, and real-world plausibility. Images scoring below 5 in any category were removed (21,808 images, 64.77\%). See the Appendix for examples of excluded low-quality images; \textbf{(3) Image Category Classification}: We next asked GPT-4o to classify the category of each image, resulting in 57 categories obtained. The most common categories were \textit{Charts and Graphs} (22.92\%) and \textit{Diagrams and Schematics} (11.17\%). Categories with fewer than 50 images were excluded (213 images, 1.80\%); (4) \textbf{Weighted Image Category Sampling}: To ensure our benchmark remains compact and accessible for other researchers to evaluate without excessive computational burden, we maintained a relatively small scale. Accordingly, we performed a weighted random sampling of images from each category, proportional to the overall category distribution, while capping the total number of selected images at 1K. 

\paragraph{Step 2: Personal Background Generation.} 

To construct more plausible and realistic multi-turn conversations, we draw inspiration from real-world scenarios. Most people typically engage in dialogue with a specific goal in mind—such as resolving a problem by providing an image alongside a relevant query—and these problems often relate to the user's situational context and personal background (\eg age, profession, hobbies). Hence, before generating the multi-turn conversation based solely on the image, we first create a \textit{fictional character} and \textit{scenario context} to represent a multi-turn exchange between this character and an AI assistant, along with a \textit{goal} tied to the scenario, using GPT-4o.\footnote{In pilot experiments, we observed that generating multi-turn conversations solely from an image led to less diverse and lower-quality outcomes.} Leveraging the \textit{personal background-to-conversation} approach~\cite{kim2022soda,jang2023conversation,lee2024stark} enables the construction of multi-turn conversation datasets with LLMs, leading to richer, more diverse, and higher-quality dialogues.

\paragraph{Step 3: Multi-turn Conversations Generation.} 

Building on the fictional character, scenario context, and defined goal, we use GPT-4o to generate multi-turn conversations between the character and the AI assistant. These conversations follow four key principles:
\begin{itemize}[leftmargin=*]
    \item \textbf{Detailed \& Informative Responses:} The assistant’s responses must be detailed, specific, expert-level, and highly informative.

    \item \textbf{Increasing Complexity:} Each subsequent user query becomes progressively more challenging, creative, and complex, as motivated by prior work~\cite{liu2024convbench}.

    \item \textbf{Diverse Linguistic Styles:} To emulate realistic human interactions, user utterances vary in linguistic style instead of relying on task-specific prompts commonly seen in existing benchmarks~\cite{liu2024mmdu} (\eg ``describe,'' ``what is,'' ``where is'').
    
    \item \textbf{Conversation Length:} Each conversation consists of four turns, yielding a total of eight utterances (four from the user and four from the assistant).
\end{itemize}
In total, we obtained 839 multi-turn conversations after removing 161 samples due to degeneracy issues with GPT-4o, such as parsing failures or insufficient turn length (\textless 6).

\paragraph{Step 4: Manual Reviewing.} 

To build a realistic and rigorous multi-turn conversation benchmark, we manually review the generated conversations according to three primary criteria:
\begin{itemize}[leftmargin=*]
    \item \textbf{Naturalness \& Realism:} We remove any conversation that does not flow realistically or contains user queries that are implausible given the image. For example, if the image is a complex scatter plot with vibrant color dots, yet the user's query solely focuses on the color rather than the actual content of the plot, we deem it unrealistic and remove the sample.

    \item \textbf{Correctness:} We conservatively remove conversations if the AI assistant's response is clearly incorrect or provides misleading information.

    \item \textbf{Blindness:} Inspired by prior works~\cite{li2024naturalbench,chen2024far} addressing the ``blind'' issue, we remove conversations in which the AI assistant can sufficiently solve the user's query without referencing the provided image.

\end{itemize}
Following this process, we retained 647 conversations and removed 192. Specifically, 48 were removed for unnatural flow, 65 for incorrect responses, and 104 for blindness-related issues, with some overlap across these criteria.

\subsection{Analysis of \textbf{\datasetName}}

\paragraph{Basic Statistics.} As shown in Table~\ref{main_tab:basic_stat}, \datasetName consists of 647 conversations, averaging 3.91 turns per conversation. It spans 8 primary interaction goals and 484 sub-interaction goals, allowing for a comprehensive evaluation of VLMs' ability to handle diverse and realistic user interactions. Additionally, \datasetName covers nine main tasks with 484 corresponding sub-tasks. \datasetName exhibits high lexical diversity in both queries and responses, suggesting that it necessitates advanced reasoning to effectively address complex user queries.

{\renewcommand{\arraystretch}{1.2}
\begin{table}[t]
\centering
\begin{adjustbox}{width=0.8\linewidth}
\begin{tabular}{lc}
 \toprule
 \textbf{Statistic} & \textbf{Number} \\
 \midrule
  Total Conversations & 647 \\
  Unique number of images & 647 \\
  Avg./Max. number of turns & 3.91/4 \\
  Interaction goal/sub-goal classes & 8/484 \\
  Image main/sub classes & 25/384 \\
  Task main/sub classes & 9/484 \\
  \midrule
  Avg./Max. question length & 30.53/70 \\
  Avg./Max. answer length & 221.51/742 \\
  \midrule
  Avg./Max. question diversity & 111.97/235.48 \\ 
  Avg./Max. answer diversity & 117.96/230.48 \\
  \midrule
  Unique number of checklist items & 21995 \\
  Unique number of main/sub key aspect & 37/10115 \\
 \bottomrule
 \end{tabular}
\end{adjustbox}
\caption{Statistics of \datasetName. Lexical diversity of questions and answers is measured using MTLD~\cite{mccarthy2010mtld}, while question and answer length is computed with the LLaMA-3.1-8B tokenizer.}
 \label{main_tab:basic_stat}
\vspace{-1em}
\end{table}}

\paragraph{Detailed Distributions.} Figure~\ref{main_fig:detailed_distribution} presents the detailed distribution of (a) tasks, (b) interaction goals, and (c) image domains in \datasetName. (a) \datasetName exhibits a high proportion of reasoning tasks, followed by knowledge-related tasks. Compared to MMDU and ConvBench, \datasetName includes more advanced reasoning tasks (\eg, mathematics, charts and diagrams, coding). (b) \datasetName highlights the diversity of interaction goals in \datasetName, where ``Analysis'' is the primary motivation for multi-turn interactions in \datasetName, followed by ``Verification'', both of which reflect realistic and practical interaction scenarios. (c) \datasetName contains a substantial number of image domains, with ``Charts and Graphs'' being the most predominant, followed by ``Diagrams and Schematics''. These distributions demonstrate \datasetName's capacity to comprehensively evaluate VLM capabilities across diverse and challenging tasks, realistic interaction goals, and a broad spectrum of image types.

\subsection{Automatic Evaluation Process}

Recently, the use of powerful LLMs/VLMs (e.g., GPT-4o) as evaluators has become a standard approach for assessing response quality in open-ended tasks~\cite{zheng2023judging,kim2023prometheus,kim2024prometheus,lee2024prometheusvision}. This is typically done by assigning an integer score between 1 and 10 to measure response quality. While this method is both straightforward and effective, recent studies~\cite{lin2024wildbench,lee2024checkeval} have raised concerns about its robustness. Therefore, checklist-based evaluation methods have been introduced to improve reliability and interpretability. In this work, we adopt an instance-specific, checklist-based evaluation approach to provide more interpretable and robust evaluation results. Specifically, we generate a unique checklist for each user query at every turn in a multi-turn conversation. Each checklist consists of multiple binary questions, with each question focusing on a specific key aspect (\eg perception, factual correctness). To ensure quality, we generate these checklists using GPT-4o and Claude-3.5-Sonnet, followed by a manual review conducted by the authors to validate their accuracy and reliability.

\begin{figure}[!t]
    \centering
    \includegraphics[width=0.65\linewidth]{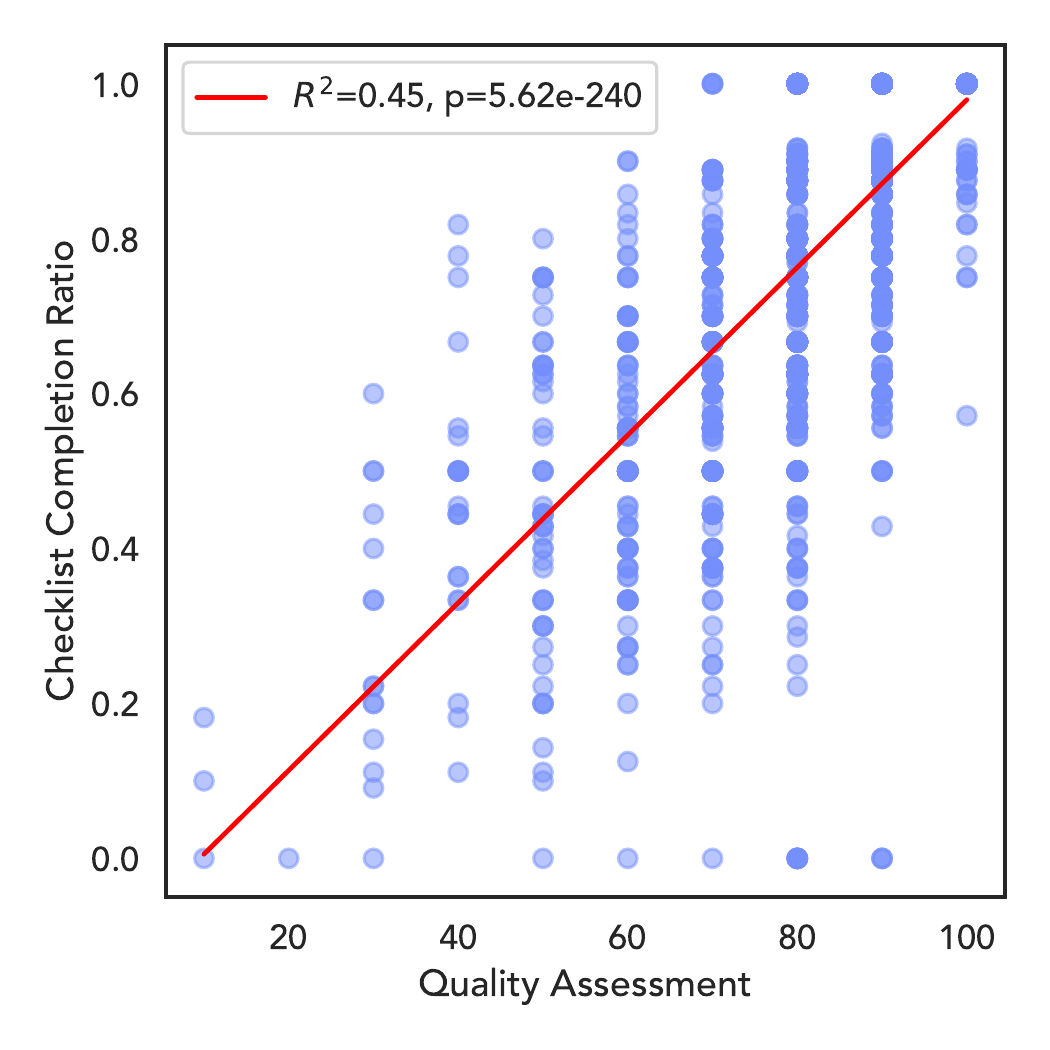}
    \vspace{-1em}
    \caption{Correlation between the checklist completion ratio and quality assessment. The red line represents the best-fit linear regression, indicating a strong positive correlation \((R^2=0.44)\) between the two metrics.}
    \label{main_fig:corr_checklist_direct}
\end{figure}

\paragraph{Evaluation Metrics.} 
We employ GPT-4o-2024-11-20 as the evaluator VLM, prompting it to assess the quality of generated responses at every turn in a multi-turn conversation based on a checklist and a reference answer~\footnote{Existing studies have demonstrated that providing a reference answer improves the correlation between human and model judgments. Therefore, we incorporate a reference answer in our evaluation.}. The prompt template used for evaluation is provided in the Appendix. Intuitively, we posit that a high-quality response should successfully satisfy the checklist items while achieving a higher integer score on a 1 to 10 scale. Based on this premise, our metric consists of two sub-metrics: 
\begin{itemize}
    \item \textbf{Checklist Completion Ratio:} This measures how well the generated response addresses the given checklist items by calculating the ratio of ``Yes''.
    \item \textbf{Quality Assessment:} This assigns an integer score between 1 and 10 to evaluate the overall quality of the response. The score is then scaled up from 1–10 to 10–100 for better interpretability.
\end{itemize}

As shown in Figure~\ref{main_fig:corr_checklist_direct}, Checklist Completion Ratio and Quality Assessment exhibit a strong positive correlation, indicating that responses meeting more checklist items tend to receive higher quality scores. Based on this, our final score is computed as their product.

\newcommand{\negativecell}{\cellcolor[HTML]{ffe3e3}}
\newcommand{\lowcell}{\cellcolor[HTML]{d0ebff}}
\newcommand{\middlecell}{\cellcolor[HTML]{a5d8ff}}
\newcommand{\highcell}{\cellcolor[HTML]{4dabf7}}

{\renewcommand{\arraystretch}{1.35}
\begin{table}[t]
\centering
\begin{adjustbox}{width=\linewidth}
\begin{tabular}{l|cccc|c|c}
\toprule
Models                 & Turn 1 & Turn 2 & Turn 3 & Turn 4 & \textit{Avg}.  & $r$ \\ \midrule
InternVL2.5-1B       & 13.93  & 21.36  & 23.51  & 26.08  & 21.22 & \middlecell 3.86  \\
InternVL2.5-2B       & 17.79  & 28.96  & 30.80  & 34.32  & 27.97 & \highcell 5.14  \\
InternVL2.5-4B       & 27.14  & 37.24  & 38.79  & 39.46  & 35.66 & \middlecell 3.85  \\
Qwen2.5-VL-3B        & 32.80  & 37.41  & 38.55  & 44.05  & 38.20 & \middlecell 3.49  \\ \midrule
LLaVA-1.5-7B         & 9.10   & 26.43  & 29.14  & 31.81  & 24.12 & \highcell 7.08  \\
LLaVA-NeXT-7B        & 13.59  & 28.25  & 29.94  & 33.07  & 26.21 & \highcell 6.01  \\
LLaVA-OneVision-7B   & 24.83  & 34.47  & 37.72  & 37.81  & 33.71 & \highcell 4.22  \\
Qwen2-VL-7B          & 35.93  & 37.99  & 39.35  & 43.10  & 39.09 & \middlecell 2.29  \\
Qwen2.5-VL-7B        & 45.13  & 47.60  & 49.31  & 50.58  & \underline{48.15} & \lowcell 1.81  \\
InternVL2-8B         & 32.09  & 39.37  & 38.98  & 38.88  & 37.33 & \middlecell 2.00  \\
InternVL2.5-8B       & 22.90  & 37.85  & 38.89  & 42.15  & 35.45 & \highcell 5.88  \\
LLaMA-3.2-11B-Vision & 14.38  & 20.02  & 25.41  & 26.63  & 21.61 & \highcell 4.21  \\
Qwen2.5-VL-72B       & 52.05  & 47.72  & 45.82  & 46.19  & 47.95 & \negativecell -1.95 \\
LLaMA-3.2-90B-Vision & 24.92  & 37.63  & 38.99  & 41.58  & 35.78 & \highcell 5.13  \\ \midrule
Gemini-2.0-Flash     & 42.03  & 49.37  & 51.23  & 48.41  & 47.76 & \middlecell 2.10  \\
Claude-3.5-Sonnet    & 46.60  & 47.16  & 48.30  & 45.00  & 46.76 & \negativecell -0.37 \\
Claude-3.7-Sonnet    & 43.38  & 44.39  & 44.77  & 43.70  & 44.06 & \lowcell 0.13  \\
GPT-4o               & 48.56  & 50.28  & 50.54  & 49.12  & \textbf{49.63} & \lowcell 0.19  \\ \midrule
\end{tabular}
\end{adjustbox}
\caption{Average performance of 18 VLMs across multiple turns in \datasetName under the \texttt{Oracle} setting. $r$ represents the slope of the performance increase as interactions progress.}
\label{main_tab:main_oracle_result}
\vspace{-1em}
\end{table}}

\section{Experiments} \label{main_sec:expr}

\subsection{Experimental Setup}
\paragraph{Evaluated Models.}
We evaluate in total 18 VLMs, categorized into: (i) Proprietary models: GPT-4o-2024-11-20~\cite{gpt4o}, Claude-3.5-Sonnet-2024-10-22~\cite{anthropic2024claude}, Claude-3.7-Sonnet-2025-02-19~\cite{claude3.7sonnet}, Gemini-2.0-Flash-001~\cite{gemini2.0flash}; (ii) Open-source models: LLaVA-1.5-7B~\cite{liu2024visual}, LLaVA-NeXT-7B~\cite{liu2024llava}, LLaVA-OneVision-7B~\cite{li2024llavaone}, Qwen2-VL-7B~\cite{Qwen2VL}, Qwen2.5-VL-\{3, 7, 72\}B-Instruct~\cite{Qwen2.5-VL}, LLaMA-3.2-\{11, 90\}B-Vision-Instruct~\cite{llama3.2vision}, InternVL2-8B~\cite{chen2023internvl}, InternVL2.5-\{1,2,4,8\}B~\cite{chen2024expanding}.

\paragraph{Evaluation Settings.}
Since \datasetName is a multi-turn conversational benchmark, the dialogue history significantly impacts response quality. To ensure a fair comparison of different VLMs, we adopt the ground-truth dialogue history provided by \datasetName as the default evaluation setting, \texttt{\textbf{Oracle}}. This approach is widely used in existing multi-turn conversation benchmarks~\cite{bai2024mt,liu2024convbench}. However, relying solely on ground-truth dialogue history may introduce biases, as variations in linguistic style and model capacity can disproportionately affect performance across different VLMs. Therefore, we additionally assess VLMs under the \texttt{\textbf{Self-Prediction}} setting, in which the model generates its own dialogue history.

\begin{figure}[t]
    \centering
    \includegraphics[width=0.9\linewidth]{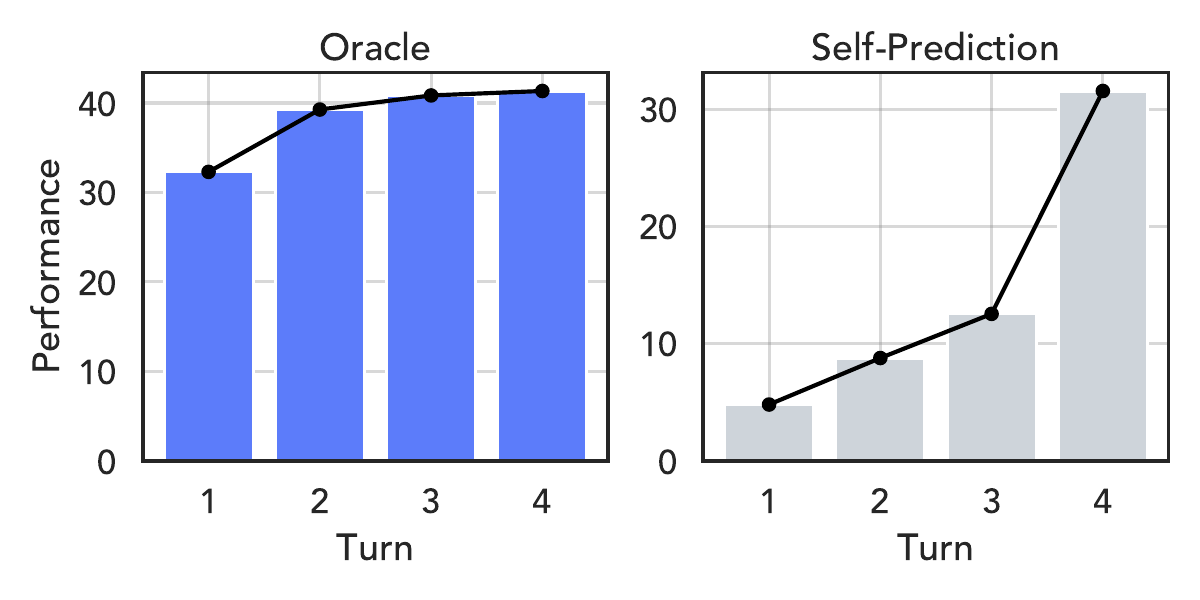}
    \vspace{-1em}
    \caption{Average performance of 18 VLMs under two different evaluation settings: \texttt{Oracle} and \texttt{Self-Prediction}.}
    \label{main_fig:turn_perf}
\end{figure}

\begin{figure}[!t]
    \centering
    \includegraphics[width=0.9\linewidth]{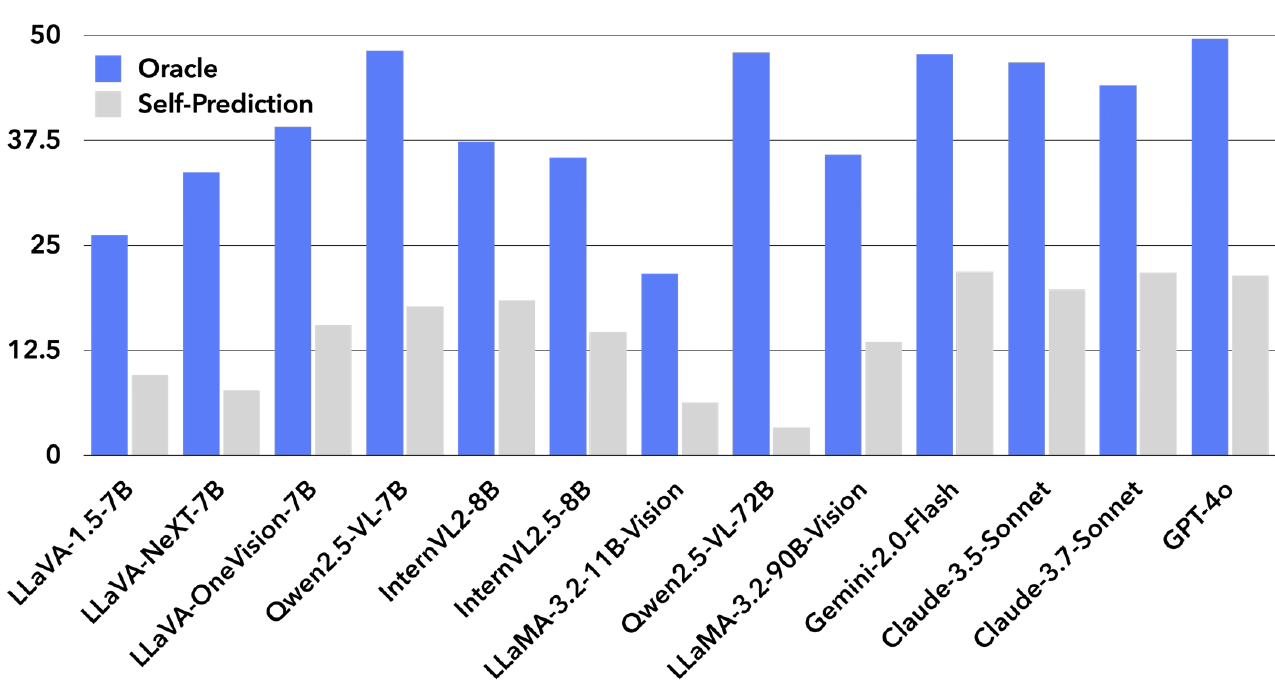}
    \vspace{-1em}
    \caption{Comparison performance across 19 different VLMs between \texttt{Oracle} and \texttt{Self-Prediction} settings.}
    \label{main_fig:oracle_vs_self}
\end{figure}

\subsection{Experimental Results and Analysis}

\paragraph{Powerful VLMs still struggle with multi-turn interactions.}
As shown in Table~\ref{main_tab:main_oracle_result}, all VLMs exhibit relatively low performance ($<50\%$), indicating that multi-turn interactions in \datasetName remain challenging, even for high-performing VLMs in the \texttt{Oracle} setting. GPT-4o achieves the highest average performance. Overall, proprietary VLMs generally outperform open-source VLMs, with the exception of the Qwen2.5-VL series, which notably surpasses Claude-3.5-Sonnet, Claude-3.7-Sonnet, and Gemini-2.0-Flash. Additionally, Claude-3.7-Sonnet scores 2.7\% lower than Claude-3.5-Sonnet. Among open-source models, LLaMA-3.2-11B-Vision achieves the lowest performance, performing even worse than LLaVA-1.5-7B.

\begin{figure*}[!t]
    \centering
    \includegraphics[width=0.95\linewidth]{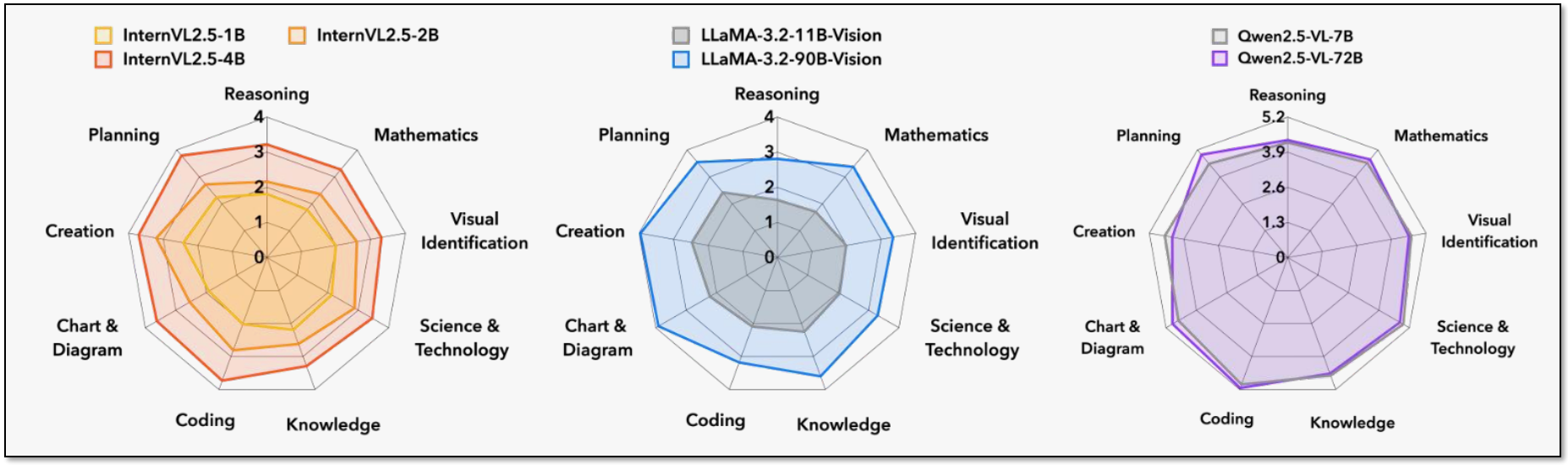}
    \caption{Performance comparison across different model sizes in three model families: InternVL2.5, LLaMA-3.2-Vision, and Qwen2.5-VL. The radar charts show that larger models generally perform better across tasks, but scaling effects vary. Notably, Qwen2.5-VL-72B excels in structured reasoning tasks (e.g., mathematics, coding), while Qwen2.5-VL-7B shows stronger creative abilities, highlighting task-dependent scaling impacts.}
    \label{main_fig:model_scaling}
\end{figure*}

\paragraph{As interaction progresses, providing ground-truth dialogue context unlocks the reasoning capabilities.}
As shown in Figure~\ref{main_fig:turn_perf}, overall performance steadily improves across multiple turns in the \texttt{Oracle} setting. This result suggests that incorporating the golden dialogue history helps VLMs better understand contextual nuances, serving as a guiding mechanism for resolving user queries in long-term conversations. By leveraging the golden dialogue, even smaller or weaker VLMs can adapt to similar linguistic styles and phrases from context, leading to improved performance. Further analysis in Table~\ref{main_tab:main_oracle_result} presents the performance improvement of each model across multi-turn interactions by measuring the slope ($r$). Overall, most VLMs benefit from extended interactions, demonstrating the effects of in-context learning. Interestingly, smaller or weaker VLMs experience the most significant gains, indicating that the ground-truth dialogue history in \datasetName helps unlock their reasoning capabilities by providing strong contextual guidance. However, Qwen2.5-VL-72B and Claude-3.5-Sonnet do not show improvement when using the ground-truth dialogue history. This negative slope may be attributed to linguistic differences between these models and GPT-4o, which serves as the reference dialogue in \datasetName.

\paragraph{Oracle \textit{vs.} Self-Prediction.} 
Figure~\ref{main_fig:oracle_vs_self} compares the performance of VLMs under \texttt{Oracle} and \texttt{Self-Prediction}. Overall, models exhibit significantly improved responses in the \texttt{Oracle} setting compared to \texttt{Self-Prediction} that relies on self-generated dialogue history, leading to a substantial performance gap (with a maximum drop of -44.64\% observed in Qwen2.5-VL-72B). These findings indicate that providing ground-truth dialogue history serves as a form of in-context learning, effectively guiding models to generate responses that are more closely aligned with reference answers. Interestingly, Qwen2.5-VL-Series models (7B and 72B) demonstrate a notably larger performance gap compared to other VLMs (-30.44 for 7B, -44.64 for 72B). This suggests that the backbone language models of Qwen2.5-VL and GPT-4o may exhibit distinct linguistic styles or response patterns. These results underscore the critical role of supplying precise and coherent dialogue context to enhance overall model performance.

\newcommand{\herelowcell}{\cellcolor[HTML]{fff9db}}
\newcommand{\heremiddlecell}{\cellcolor[HTML]{ffec99}}
\newcommand{\herehighcell}{\cellcolor[HTML]{ffd43b}}

{\renewcommand{\arraystretch}{1.35}
\begin{table}[t]
\centering
\begin{adjustbox}{width=\linewidth}
\begin{tabular}{l|cccccccc}
\toprule
Models               & \verification & \analysis & \exploration & \optimization & \calculation & \understanding & \research & \creation  \\ \midrule
InternVL2.5-1B       & \herelowcell 20.37 & \herelowcell 24.55 & \herelowcell 23.98 & \herelowcell 15.06 & \herelowcell 16.95 & \herelowcell 22.03 & \herelowcell 18.36 & \herelowcell 22.34  \\
InternVL2.5-2B       & \herelowcell 26.67 & \herelowcell 31.48 & \herelowcell 24.82 & \herelowcell 22.74 & \herelowcell 26.23 & \herelowcell 28.70 & \herelowcell 26.51 & \herelowcell 27.38  \\
InternVL2.5-4B       & \heremiddlecell 34.43 & \herehighcell 38.56 & \heremiddlecell 32.14 & \herelowcell 31.56 & \heremiddlecell 34.46 & \herehighcell 40.90 & \herelowcell 29.65 & \heremiddlecell 35.17  \\
Qwen2.5-VL-3B        & \heremiddlecell 36.08 & \herehighcell 43.19 & \heremiddlecell 34.56 & \herelowcell 29.43 & \heremiddlecell 37.61 & \herehighcell 43.46 & \herelowcell 32.21 & \heremiddlecell 35.91 \\ \midrule
LLaVA-1.5-7B         & \herelowcell 20.34 & \herelowcell 26.58 & \herelowcell 26.27 & \herelowcell 19.57 & \herelowcell 14.11 & \herelowcell 23.32 & \herelowcell 28.70 & \herelowcell 27.59  \\
LLaVA-NeXT-7B        & \herelowcell 21.78 & \herelowcell 30.36 & \herelowcell 28.62 & \herelowcell 21.75 & \herelowcell 14.26 & \herelowcell 25.41 & \herelowcell 29.06 & \herelowcell 29.16  \\
LLaVA-OneVision-7B   & \heremiddlecell 30.55 & \heremiddlecell 36.29 & \heremiddlecell 31.58 & \herelowcell 27.97 & \heremiddlecell 31.84 & \heremiddlecell 37.27 & \heremiddlecell 31.73 & \heremiddlecell 36.14 \\
Qwen2-VL-7B          & \heremiddlecell 36.28 & \herehighcell 43.66 & \heremiddlecell 35.95 & \heremiddlecell 31.62 & \heremiddlecell 35.77 & \herehighcell 43.54 & \heremiddlecell 35.79 & \heremiddlecell 37.13  \\
Qwen2.5-VL-7B        & \herehighcell 46.14 & \herehighcell 54.23 & \herehighcell 44.23 & \herehighcell 40.64 & \herehighcell 44.86 & \herehighcell 54.71 & \herehighcell 39.73 & \herehighcell 44.45  \\
InternVL2-8B         & \heremiddlecell 34.85 & \herehighcell 42.06 & \heremiddlecell 33.00 & \herelowcell 28.76 & \heremiddlecell 33.74 & \herehighcell 41.76 & \heremiddlecell 33.19 & \heremiddlecell 37.96 \\
InternVL2.5-8B       & \heremiddlecell 33.55 & \heremiddlecell 38.39 & \heremiddlecell 32.95 & \herelowcell 29.92 & \heremiddlecell 34.45 & \herehighcell 39.58 & \herelowcell 30.64 & \heremiddlecell 36.98 \\
LLaMA-3.2-11B-Vision & \herelowcell 17.94 & \herelowcell 23.30 & \herelowcell 20.43 & \herelowcell 18.84 & \herelowcell 16.51 & \herelowcell 22.48 & \herelowcell 23.41 & \herelowcell 26.03  \\
Qwen2.5-VL-72B       & \herehighcell 45.58 & \herehighcell 52.10 & \herehighcell 42.86 & \herehighcell 42.59 & \herehighcell 47.15 & \herehighcell 55.40 & \herehighcell 43.32 & \herehighcell 39.82  \\
LLaMA-3.2-90B-Vision & \heremiddlecell 33.96 & \heremiddlecell 38.19 & \heremiddlecell 33.41 & \heremiddlecell 32.17 & \heremiddlecell 34.68 & \heremiddlecell 39.03 & \heremiddlecell 32.57 & \heremiddlecell 35.84  \\
\midrule
Gemini-2.0-Flash     & \herehighcell 43.68 & \herehighcell 52.25 & \herehighcell 40.73 & \herehighcell 44.46 & \herehighcell 48.92 & \herehighcell 54.48 & \herehighcell 42.58 & \herehighcell 42.17  \\
Claude-3.5-Sonnet    & \herehighcell 46.15 & \herehighcell 50.94 & \herehighcell 43.80 & \herehighcell 40.43 & \herehighcell 47.72 & \herehighcell 53.95 & \herehighcell 37.99 & \herehighcell 41.08  \\
Claude-3.7-Sonnet    & \herehighcell 42.58 & \herehighcell 47.43 & \herehighcell 43.18 & \herehighcell 38.83 & \herehighcell 45.33 & \herehighcell 51.22 & \herehighcell 38.56 & \heremiddlecell 34.65  \\
GPT-4o               & \herehighcell 46.80 & \herehighcell 53.67 & \herehighcell 42.70 & \herehighcell 46.36 & \herehighcell 50.54 & \herehighcell 56.41 & \herehighcell 43.50 & \herehighcell 44.51 \\ \bottomrule


\end{tabular}
\end{adjustbox}
\caption{Performance of 18 VLMs in \datasetName for 8 different interaction goals---verification (\verification), analysis (\analysis), exploration (\exploration), optimization (\optimization), calculation (\calculation), understanding (\understanding), research (\research), and creation (\creation)---under \texttt{Oracle} setting.}
\label{main_tab:main_goal_result}
\vspace{-1em}
\end{table}}

\begin{figure*}[!t]
    \centering
    \includegraphics[width=0.85\linewidth]{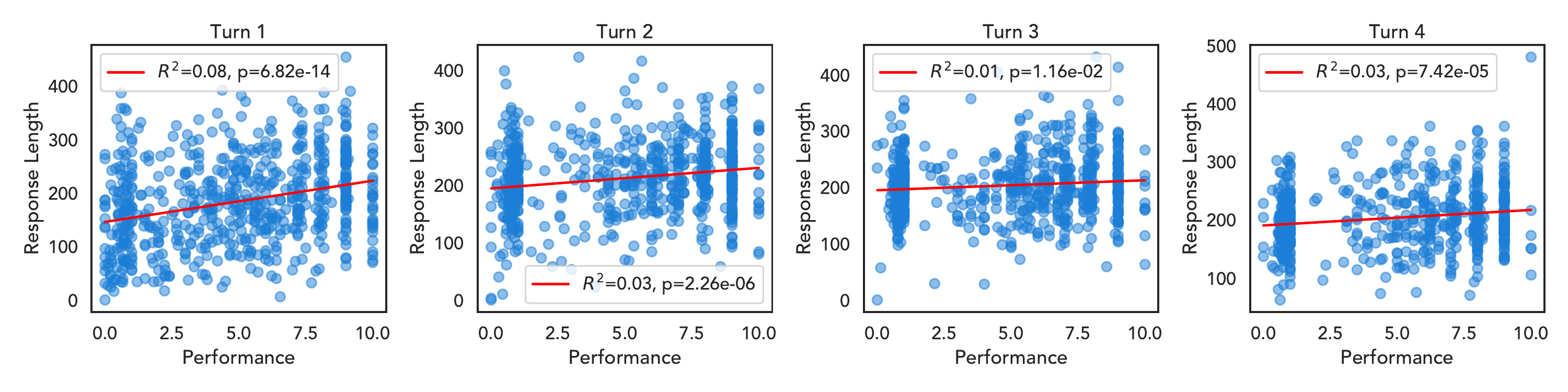}
    \caption{Linear correlation ($R^2$) between response length and performance from GPT-4o.}
    \label{main_fig:verbosity_bias}
    \vspace{-1em}
\end{figure*}

\paragraph{VLMs demonstrate selective interactivity capabilities.} As shown in Table~\ref{main_tab:main_goal_result}, most VLMs exhibit strong performance in analysis (\analysis) and understanding (\understanding). However, their performance in optimization (\optimization) and research (\research) tasks—typically demanding innovative thinking—remains comparatively weak. Among these, the Qwen2.5-VL series demonstrates relatively stronger interactivity across all interaction goals. On the other hand, weaker or smaller VLMs with lower performance, such as LLaVA-1.5-7B and LLaVA-NeXT-7B, struggle particularly with verifiable tasks, including optimization (\optimization) and calculation (\calculation).

\paragraph{Effect of Model Scaling.} Figure~\ref{main_fig:model_scaling} presents performance comparisons across varying model sizes in three different VLM groups: InternVL2.5, LLaMA-3.2-Vision, and Qwen2.5-VL. In general, larger models exhibit higher overall performance on most tasks. However, within the Qwen2.5-VL series, increasing the model size does not always guarantee improvements, and its impact appears task-dependent. For instance, Qwen2.5-VL-72B achieves stronger reasoning on verifiable tasks (\eg mathematics, coding, and interpreting charts \& diagrams), while Qwen2.5-VL-7B shows advantages in creative thinking. These findings suggest that larger models generally offer enhanced interactivity capabilities, although the benefit may vary based on the specific task.

\paragraph{Does automatic evaluation metric possess verbosity bias?} We explore whether our evaluation metric favors longer answers across multi-turns. As shown in Figure~\ref{main_fig:verbosity_bias}, the linear correlation ($R^2$) between length of response and performance from GPT-4o is weaken as interaction increases (until turn 3). As interaction progress until turn 3, the $p$-value increases and $R$-value decreases, which suggests that our evaluation metric could reduce verbosity bias and length dependency.

\section{Related Work}

\paragraph{Single-Turn Evaluation Benchmarks.} 
To comprehensively evaluate multiple vision-language capabilities of recent powerful VLMs, many benchmarks have been introduced. These benchmarks assess integrated capabilities (\ie perception and reasoning)~\cite{chen2024we,fu2023mme,zhang2024mme,yue2024mmmu,yue2024mmmupro,pi2024image,yu2023mm,yu2024mm,liu2023mmbench,li2023seed,li2024seed,wu2023q}, text-richness (\eg charts, diagrams)~\cite{kembhavi2016diagram,masry2022chartqa}, mathematical reasoning~\cite{lu2023mathvista,kamath2023s,zhang2024mathverse}, scientific understanding~\cite{lu2022learn}, perception-specialized tasks that are easy for humans to perceive~\cite{zhang2025vlm2,fu2024blink,tong2024eyes,li2024naturalbench,rahmanzadehgervi2024vision,tong2024cambrian, pi2024personalized}, and web \& mobile applications~\cite{liu2024visualwebbench}. Despite their breadth, most existing benchmarks rely on single-turn evaluations—typically in the form of binary, multiple-choice, or free-form responses—thus failing to capture the effectiveness of VLMs in multi-turn interaction. This gap highlights the need for a new benchmark that better reflects practical, real-world conversational scenarios.

\paragraph{Multi-Turn Evaluation Benchmarks.}
In the NLP domain, numerous multi-turn conversation benchmarks have recently been introduced to assess the interactive capabilities of LLMs~\cite{wang2023mint,zheng2023judging,bai2024mt,kwan2024mt,he2024multi,sirdeshmukh2025multichallenge,lin2024wildbench}. By contrast, only a few benchmarks exist for evaluating the multi-turn interaction capabilities of VLMs, such as MMDU \cite{liu2024mmdu} and ConvBench \cite{liu2024convbench}. However, these existing benchmarks still fall short of assessing VLM performance in real-world human-AI interactions. In particular, they do not address more advanced and complex reasoning tasks (\eg verifying equations shown in the image), and most queries can be resolved through straightforward statements rather than more elaborate formats, such as code or mathematical expressions. These limitations hinder a comprehensive evaluation of VLMs under a wide range of scenarios, underscoring the need for a richer, more rigorous multi-turn conversation benchmark.

\section{Conclusion}
In this work, we introduce \datasetName, a new multi-turn conversation benchmark designed to evaluate VLMs. Our dataset encompasses a diverse range of interaction goals (\eg verification, analysis) and tasks (\eg reasoning, mathematics), providing a realistic and challenging testbed for assessing the multi-turn interactive capabilities of VLMs. Moreover, we propose an instance-specific, checklist-based evaluation framework comprising two key sub-metrics: checklist completion ratio and quality assessment. We reveal that even recent state-of-the-art VLMs struggle with handling multi-turn interactions effectively. 

\section*{Acknowledgement}

This work was partly supported by Institute of Information \& communications Technology Planning \& Evaluation (IITP) grant funded by the Korea government(MSIT) (RS-2022-II220641, XVoice: Multi-Modal Voice Meta Learning)

{
    \small
    \bibliographystyle{ieeenat_fullname}
    \bibliography{main}

\begin{thebibliography}{79}
\providecommand{\natexlab}[1]{#1}
\providecommand{\url}[1]{\texttt{#1}}
\expandafter\ifx\csname urlstyle\endcsname\relax
  \providecommand{\doi}[1]{doi: #1}\else
  \providecommand{\doi}{doi: \begingroup \urlstyle{rm}\Url}\fi

\bibitem[{Anthropic}(2024)]{claude3series2024}
{Anthropic}.
\newblock The claude 3 model family: Opus, sonnet, haiku.
\newblock \url{https://www.anthropic.com}, 2024.

\bibitem[Anthropic(2025)]{claude3.7sonnet}
Anthropic.
\newblock Claude 3.7 sonnet system card.
\newblock 2025.

\bibitem[Anthropic(2024)]{anthropic2024claude}
AI Anthropic.
\newblock Claude 3.5 sonnet model card addendum.
\newblock \emph{Claude-3.5 Model Card}, 3\penalty0 (6), 2024.

\bibitem[Bai et~al.(2024)Bai, Liu, Bu, He, Liu, Zhou, Lin, Su, Ge, Zheng, et~al.]{bai2024mt}
Ge Bai, Jie Liu, Xingyuan Bu, Yancheng He, Jiaheng Liu, Zhanhui Zhou, Zhuoran Lin, Wenbo Su, Tiezheng Ge, Bo Zheng, et~al.
\newblock Mt-bench-101: A fine-grained benchmark for evaluating large language models in multi-turn dialogues.
\newblock \emph{arXiv preprint arXiv:2402.14762}, 2024.

\bibitem[Bai et~al.(2025)Bai, Chen, Liu, Wang, Ge, Song, Dang, Wang, Wang, Tang, Zhong, Zhu, Yang, Li, Wan, Wang, Ding, Fu, Xu, Ye, Zhang, Xie, Cheng, Zhang, Yang, Xu, and Lin]{Qwen2.5-VL}
Shuai Bai, Keqin Chen, Xuejing Liu, Jialin Wang, Wenbin Ge, Sibo Song, Kai Dang, Peng Wang, Shijie Wang, Jun Tang, Humen Zhong, Yuanzhi Zhu, Mingkun Yang, Zhaohai Li, Jianqiang Wan, Pengfei Wang, Wei Ding, Zheren Fu, Yiheng Xu, Jiabo Ye, Xi Zhang, Tianbao Xie, Zesen Cheng, Hang Zhang, Zhibo Yang, Haiyang Xu, and Junyang Lin.
\newblock Qwen2.5-vl technical report.
\newblock \emph{arXiv preprint arXiv:2502.13923}, 2025.

\bibitem[Bitton et~al.(2023)Bitton, Bansal, Hessel, Shao, Zhu, Awadalla, Gardner, Taori, and Schmidt]{bitton2023visit}
Yonatan Bitton, Hritik Bansal, Jack Hessel, Rulin Shao, Wanrong Zhu, Anas Awadalla, Josh Gardner, Rohan Taori, and Ludwig Schmidt.
\newblock Visit-bench: A benchmark for vision-language instruction following inspired by real-world use.
\newblock \emph{arXiv preprint arXiv:2308.06595}, 2023.

\bibitem[Cha et~al.(2024)Cha, Kang, Mun, and Roh]{cha2024honeybee}
Junbum Cha, Wooyoung Kang, Jonghwan Mun, and Byungseok Roh.
\newblock Honeybee: Locality-enhanced projector for multimodal llm.
\newblock In \emph{Proceedings of the IEEE/CVF Conference on Computer Vision and Pattern Recognition}, pages 13817--13827, 2024.

\bibitem[Chen et~al.(2024{\natexlab{a}})Chen, Liang, Siu, Wang, Wang, Wang, Ni, Zhu, Jiang, Lyu, et~al.]{chen2024mega}
Jiacheng Chen, Tianhao Liang, Sherman Siu, Zhengqing Wang, Kai Wang, Yubo Wang, Yuansheng Ni, Wang Zhu, Ziyan Jiang, Bohan Lyu, et~al.
\newblock Mega-bench: Scaling multimodal evaluation to over 500 real-world tasks.
\newblock \emph{arXiv preprint arXiv:2410.10563}, 2024{\natexlab{a}}.

\bibitem[Chen et~al.(2024{\natexlab{b}})Chen, Li, Dong, Zhang, Zang, Chen, Duan, Wang, Qiao, Lin, et~al.]{chen2024we}
Lin Chen, Jinsong Li, Xiaoyi Dong, Pan Zhang, Yuhang Zang, Zehui Chen, Haodong Duan, Jiaqi Wang, Yu Qiao, Dahua Lin, et~al.
\newblock Are we on the right way for evaluating large vision-language models?
\newblock \emph{arXiv preprint arXiv:2403.20330}, 2024{\natexlab{b}}.

\bibitem[Chen et~al.(2023)Chen, Wu, Wang, Su, Chen, Xing, Zhong, Zhang, Zhu, Lu, Li, Luo, Lu, Qiao, and Dai]{chen2023internvl}
Zhe Chen, Jiannan Wu, Wenhai Wang, Weijie Su, Guo Chen, Sen Xing, Muyan Zhong, Qinglong Zhang, Xizhou Zhu, Lewei Lu, Bin Li, Ping Luo, Tong Lu, Yu Qiao, and Jifeng Dai.
\newblock Internvl: Scaling up vision foundation models and aligning for generic visual-linguistic tasks.
\newblock \emph{arXiv preprint arXiv:2312.14238}, 2023.

\bibitem[Chen et~al.(2024{\natexlab{c}})Chen, Wang, Cao, Liu, Gao, Cui, Zhu, Ye, Tian, Liu, et~al.]{chen2024expanding}
Zhe Chen, Weiyun Wang, Yue Cao, Yangzhou Liu, Zhangwei Gao, Erfei Cui, Jinguo Zhu, Shenglong Ye, Hao Tian, Zhaoyang Liu, et~al.
\newblock Expanding performance boundaries of open-source multimodal models with model, data, and test-time scaling.
\newblock \emph{arXiv preprint arXiv:2412.05271}, 2024{\natexlab{c}}.

\bibitem[Chen et~al.(2024{\natexlab{d}})Chen, Wang, Tian, Ye, Gao, Cui, Tong, Hu, Luo, Ma, et~al.]{chen2024far}
Zhe Chen, Weiyun Wang, Hao Tian, Shenglong Ye, Zhangwei Gao, Erfei Cui, Wenwen Tong, Kongzhi Hu, Jiapeng Luo, Zheng Ma, et~al.
\newblock How far are we to gpt-4v? closing the gap to commercial multimodal models with open-source suites.
\newblock \emph{arXiv preprint arXiv:2404.16821}, 2024{\natexlab{d}}.

\bibitem[Erfei~Cui(2024)]{sharegpt4o}
et~al. Erfei~Cui.
\newblock Sharegpt-4o: Comprehensive multimodal annotations with gpt-4o, 2024.
\newblock \url{https://sharegpt4o.github.io/}.

\bibitem[Fu et~al.(2023)Fu, Chen, Shen, Qin, Zhang, Lin, Yang, Zheng, Li, Sun, et~al.]{fu2023mme}
Chaoyou Fu, Peixian Chen, Yunhang Shen, Yulei Qin, Mengdan Zhang, Xu Lin, Jinrui Yang, Xiawu Zheng, Ke Li, Xing Sun, et~al.
\newblock Mme: A comprehensive evaluation benchmark for multimodal large language models.
\newblock \emph{arXiv preprint arXiv:2306.13394}, 2023.

\bibitem[Fu et~al.(2024)Fu, Hu, Li, Feng, Wang, Lin, Roth, Smith, Ma, and Krishna]{fu2024blink}
Xingyu Fu, Yushi Hu, Bangzheng Li, Yu Feng, Haoyu Wang, Xudong Lin, Dan Roth, Noah~A Smith, Wei-Chiu Ma, and Ranjay Krishna.
\newblock Blink: Multimodal large language models can see but not perceive.
\newblock \emph{arXiv preprint arXiv:2404.12390}, 2024.

\bibitem[Gemini(2024)]{gemini2.0flash}
Gemini.
\newblock Gemini 2.0 flash.
\newblock 2024.

\bibitem[He et~al.(2024)He, Jin, Wang, Bi, Mandyam, Zhang, Zhu, Li, Xu, Lv, et~al.]{he2024multi}
Yun He, Di Jin, Chaoqi Wang, Chloe Bi, Karishma Mandyam, Hejia Zhang, Chen Zhu, Ning Li, Tengyu Xu, Hongjiang Lv, et~al.
\newblock Multi-if: Benchmarking llms on multi-turn and multilingual instructions following.
\newblock \emph{arXiv preprint arXiv:2410.15553}, 2024.

\bibitem[Hurst et~al.(2024)Hurst, Lerer, Goucher, Perelman, Ramesh, Clark, Ostrow, Welihinda, Hayes, Radford, et~al.]{gpt4o}
Aaron Hurst, Adam Lerer, Adam~P Goucher, Adam Perelman, Aditya Ramesh, Aidan Clark, AJ Ostrow, Akila Welihinda, Alan Hayes, Alec Radford, et~al.
\newblock Gpt-4o system card.
\newblock \emph{arXiv preprint arXiv:2410.21276}, 2024.

\bibitem[Jang et~al.(2023)Jang, Boo, and Kim]{jang2023conversation}
Jihyoung Jang, Minseong Boo, and Hyounghun Kim.
\newblock Conversation chronicles: Towards diverse temporal and relational dynamics in multi-session conversations.
\newblock \emph{arXiv preprint arXiv:2310.13420}, 2023.

\bibitem[Jiao et~al.(2024)Jiao, Chen, Huang, Li, and Shen]{jiao2024enhancing}
Qirui Jiao, Daoyuan Chen, Yilun Huang, Yaliang Li, and Ying Shen.
\newblock Enhancing multimodal large language models with vision detection models: An empirical study.
\newblock \emph{arXiv preprint arXiv:2401.17981}, 2024.

\bibitem[Kamath et~al.(2023)Kamath, Hessel, and Chang]{kamath2023s}
Amita Kamath, Jack Hessel, and Kai-Wei Chang.
\newblock What's" up" with vision-language models? investigating their struggle with spatial reasoning.
\newblock \emph{arXiv preprint arXiv:2310.19785}, 2023.

\bibitem[Kembhavi et~al.(2016)Kembhavi, Salvato, Kolve, Seo, Hajishirzi, and Farhadi]{kembhavi2016diagram}
Aniruddha Kembhavi, Mike Salvato, Eric Kolve, Minjoon Seo, Hannaneh Hajishirzi, and Ali Farhadi.
\newblock A diagram is worth a dozen images.
\newblock In \emph{Computer Vision--ECCV 2016: 14th European Conference, Amsterdam, The Netherlands, October 11--14, 2016, Proceedings, Part IV 14}, pages 235--251. Springer, 2016.

\bibitem[Kim et~al.(2022)Kim, Hessel, Jiang, West, Lu, Yu, Zhou, Bras, Alikhani, Kim, et~al.]{kim2022soda}
Hyunwoo Kim, Jack Hessel, Liwei Jiang, Peter West, Ximing Lu, Youngjae Yu, Pei Zhou, Ronan~Le Bras, Malihe Alikhani, Gunhee Kim, et~al.
\newblock Soda: Million-scale dialogue distillation with social commonsense contextualization.
\newblock \emph{arXiv preprint arXiv:2212.10465}, 2022.

\bibitem[Kim et~al.(2023)Kim, Shin, Cho, Jang, Longpre, Lee, Yun, Shin, Kim, Thorne, et~al.]{kim2023prometheus}
Seungone Kim, Jamin Shin, Yejin Cho, Joel Jang, Shayne Longpre, Hwaran Lee, Sangdoo Yun, Seongjin Shin, Sungdong Kim, James Thorne, et~al.
\newblock Prometheus: Inducing fine-grained evaluation capability in language models.
\newblock \emph{arXiv preprint arXiv:2310.08491}, 2023.

\bibitem[Kim et~al.(2024)Kim, Suk, Longpre, Lin, Shin, Welleck, Neubig, Lee, Lee, and Seo]{kim2024prometheus}
Seungone Kim, Juyoung Suk, Shayne Longpre, Bill~Yuchen Lin, Jamin Shin, Sean Welleck, Graham Neubig, Moontae Lee, Kyungjae Lee, and Minjoon Seo.
\newblock Prometheus 2: An open source language model specialized in evaluating other language models, 2024.

\bibitem[Kwan et~al.(2024)Kwan, Zeng, Jiang, Wang, Li, Shang, Jiang, Liu, and Wong]{kwan2024mt}
Wai-Chung Kwan, Xingshan Zeng, Yuxin Jiang, Yufei Wang, Liangyou Li, Lifeng Shang, Xin Jiang, Qun Liu, and Kam-Fai Wong.
\newblock Mt-eval: A multi-turn capabilities evaluation benchmark for large language models.
\newblock \emph{arXiv preprint arXiv:2401.16745}, 2024.

\bibitem[Lai et~al.(2024)Lai, Tian, Chen, Li, Yuan, Liu, and Jia]{lai2024lisa}
Xin Lai, Zhuotao Tian, Yukang Chen, Yanwei Li, Yuhui Yuan, Shu Liu, and Jiaya Jia.
\newblock Lisa: Reasoning segmentation via large language model.
\newblock In \emph{Proceedings of the IEEE/CVF Conference on Computer Vision and Pattern Recognition}, pages 9579--9589, 2024.

\bibitem[Lee et~al.(2024{\natexlab{a}})Lee, Chung, Kim, Park, and Ro]{lee2024trol}
Byung-Kwan Lee, Sangyun Chung, Chae~Won Kim, Beomchan Park, and Yong~Man Ro.
\newblock Trol: Traversal of layers for large language and vision models.
\newblock \emph{arXiv preprint arXiv:2406.12246}, 2024{\natexlab{a}}.

\bibitem[Lee et~al.(2024{\natexlab{b}})Lee, Kim, Park, and Ro]{lee2024meteor}
Byung-Kwan Lee, Chae~Won Kim, Beomchan Park, and Yong~Man Ro.
\newblock Meteor: Mamba-based traversal of rationale for large language and vision models.
\newblock \emph{arXiv preprint arXiv:2405.15574}, 2024{\natexlab{b}}.

\bibitem[Lee et~al.(2024{\natexlab{c}})Lee, Park, Kim, and Ro]{lee2024collavo}
Byung-Kwan Lee, Beomchan Park, Chae~Won Kim, and Yong~Man Ro.
\newblock Collavo: Crayon large language and vision model.
\newblock \emph{arXiv preprint arXiv:2402.11248}, 2024{\natexlab{c}}.

\bibitem[Lee et~al.(2024{\natexlab{d}})Lee, Park, Kim, and Ro]{lee2024moai}
Byung-Kwan Lee, Beomchan Park, Chae~Won Kim, and Yong~Man Ro.
\newblock Moai: Mixture of all intelligence for large language and vision models.
\newblock \emph{arXiv preprint arXiv:2403.07508}, 2024{\natexlab{d}}.

\bibitem[Lee et~al.(2024{\natexlab{e}})Lee, Kim, Park, Kim, and Seo]{lee2024prometheusvision}
Seongyun Lee, Seungone Kim, Sue~Hyun Park, Geewook Kim, and Minjoon Seo.
\newblock Prometheus-vision: Vision-language model as a judge for fine-grained evaluation, 2024{\natexlab{e}}.

\bibitem[Lee et~al.(2024{\natexlab{f}})Lee, Kim, Kim, Cho, and Kang]{lee2024checkeval}
Yukyung Lee, Joonghoon Kim, Jaehee Kim, Hyowon Cho, and Pilsung Kang.
\newblock Checkeval: Robust evaluation framework using large language model via checklist.
\newblock \emph{arXiv preprint arXiv:2403.18771}, 2024{\natexlab{f}}.

\bibitem[Lee et~al.(2024{\natexlab{g}})Lee, Lee, Youn, Oh, Ko, Hyeon, and Choi]{lee2024stark}
Young-Jun Lee, Dokyong Lee, Junyoung Youn, Kyeongjin Oh, Byungsoo Ko, Jonghwan Hyeon, and Ho-Jin Choi.
\newblock Stark: Social long-term multi-modal conversation with persona commonsense knowledge.
\newblock \emph{arXiv preprint arXiv:2407.03958}, 2024{\natexlab{g}}.

\bibitem[Li et~al.(2023)Li, Wang, Wang, Ge, Ge, and Shan]{li2023seed}
Bohao Li, Rui Wang, Guangzhi Wang, Yuying Ge, Yixiao Ge, and Ying Shan.
\newblock Seed-bench: Benchmarking multimodal llms with generative comprehension.
\newblock \emph{arXiv preprint arXiv:2307.16125}, 2023.

\bibitem[Li et~al.(2024{\natexlab{a}})Li, Ge, Chen, Ge, Zhang, and Shan]{li2024seed}
Bohao Li, Yuying Ge, Yi Chen, Yixiao Ge, Ruimao Zhang, and Ying Shan.
\newblock Seed-bench-2-plus: Benchmarking multimodal large language models with text-rich visual comprehension.
\newblock \emph{arXiv preprint arXiv:2404.16790}, 2024{\natexlab{a}}.

\bibitem[Li et~al.(2024{\natexlab{b}})Li, Lin, Peng, Nyandwi, Jiang, Ma, Khanuja, Krishna, Neubig, and Ramanan]{li2024naturalbench}
Baiqi Li, Zhiqiu Lin, Wenxuan Peng, Jean de~Dieu Nyandwi, Daniel Jiang, Zixian Ma, Simran Khanuja, Ranjay Krishna, Graham Neubig, and Deva Ramanan.
\newblock Naturalbench: Evaluating vision-language models on natural adversarial samples.
\newblock \emph{arXiv preprint arXiv:2410.14669}, 2024{\natexlab{b}}.

\bibitem[Li et~al.(2024{\natexlab{c}})Li, Zhang, Guo, Zhang, Li, Zhang, Zhang, Li, Liu, and Li]{li2024llavaone}
Bo Li, Yuanhan Zhang, Dong Guo, Renrui Zhang, Feng Li, Hao Zhang, Kaichen Zhang, Yanwei Li, Ziwei Liu, and Chunyuan Li.
\newblock Llava-onevision: Easy visual task transfer.
\newblock \emph{arXiv preprint arXiv:2408.03326}, 2024{\natexlab{c}}.

\bibitem[Li et~al.(2024{\natexlab{d}})Li, Zhang, Zhang, Zhang, Li, Li, Ma, and Li]{li2024llava}
Feng Li, Renrui Zhang, Hao Zhang, Yuanhan Zhang, Bo Li, Wei Li, Zejun Ma, and Chunyuan Li.
\newblock Llava-next-interleave: Tackling multi-image, video, and 3d in large multimodal models.
\newblock \emph{arXiv preprint arXiv:2407.07895}, 2024{\natexlab{d}}.

\bibitem[Li et~al.(2024{\natexlab{e}})Li, Zhang, Wang, Zhong, Chen, Chu, Liu, and Jia]{li2024mini}
Yanwei Li, Yuechen Zhang, Chengyao Wang, Zhisheng Zhong, Yixin Chen, Ruihang Chu, Shaoteng Liu, and Jiaya Jia.
\newblock Mini-gemini: Mining the potential of multi-modality vision language models.
\newblock \emph{arXiv preprint arXiv:2403.18814}, 2024{\natexlab{e}}.

\bibitem[Lin et~al.(2024)Lin, Deng, Chandu, Brahman, Ravichander, Pyatkin, Dziri, Bras, and Choi]{lin2024wildbench}
Bill~Yuchen Lin, Yuntian Deng, Khyathi Chandu, Faeze Brahman, Abhilasha Ravichander, Valentina Pyatkin, Nouha Dziri, Ronan~Le Bras, and Yejin Choi.
\newblock Wildbench: Benchmarking llms with challenging tasks from real users in the wild.
\newblock \emph{arXiv preprint arXiv:2406.04770}, 2024.

\bibitem[Liu et~al.(2024{\natexlab{a}})Liu, Li, Li, Li, Zhang, Shen, and Lee]{liu2024llava}
Haotian Liu, Chunyuan Li, Yuheng Li, Bo Li, Yuanhan Zhang, Sheng Shen, and Yong~Jae Lee.
\newblock Llava-next: Improved reasoning, ocr, and world knowledge, 2024{\natexlab{a}}.

\bibitem[Liu et~al.(2024{\natexlab{b}})Liu, Li, Wu, and Lee]{liu2024visual}
Haotian Liu, Chunyuan Li, Qingyang Wu, and Yong~Jae Lee.
\newblock Visual instruction tuning.
\newblock \emph{Advances in neural information processing systems}, 36, 2024{\natexlab{b}}.

\bibitem[Liu et~al.(2024{\natexlab{c}})Liu, Song, Lin, Lam, Neubig, Li, and Yue]{liu2024visualwebbench}
Junpeng Liu, Yifan Song, Bill~Yuchen Lin, Wai Lam, Graham Neubig, Yuanzhi Li, and Xiang Yue.
\newblock Visualwebbench: How far have multimodal llms evolved in web page understanding and grounding?
\newblock \emph{arXiv preprint arXiv:2404.05955}, 2024{\natexlab{c}}.

\bibitem[Liu et~al.(2024{\natexlab{d}})Liu, Ying, Zhang, Yang, Lin, Zhang, Li, Qiao, Luo, Shao, et~al.]{liu2024convbench}
Shuo Liu, Kaining Ying, Hao Zhang, Yue Yang, Yuqi Lin, Tianle Zhang, Chuanhao Li, Yu Qiao, Ping Luo, Wenqi Shao, et~al.
\newblock Convbench: A multi-turn conversation evaluation benchmark with hierarchical capability for large vision-language models.
\newblock \emph{arXiv preprint arXiv:2403.20194}, 2024{\natexlab{d}}.

\bibitem[Liu et~al.(2023)Liu, Duan, Zhang, Li, Zhang, Zhao, Yuan, Wang, He, Liu, et~al.]{liu2023mmbench}
Yuan Liu, Haodong Duan, Yuanhan Zhang, Bo Li, Songyang Zhang, Wangbo Zhao, Yike Yuan, Jiaqi Wang, Conghui He, Ziwei Liu, et~al.
\newblock Mmbench: Is your multi-modal model an all-around player?
\newblock \emph{arXiv preprint arXiv:2307.06281}, 2023.

\bibitem[Liu et~al.(2024{\natexlab{e}})Liu, Chu, Zang, Wei, Dong, Zhang, Liang, Xiong, Qiao, Lin, et~al.]{liu2024mmdu}
Ziyu Liu, Tao Chu, Yuhang Zang, Xilin Wei, Xiaoyi Dong, Pan Zhang, Zijian Liang, Yuanjun Xiong, Yu Qiao, Dahua Lin, et~al.
\newblock Mmdu: A multi-turn multi-image dialog understanding benchmark and instruction-tuning dataset for lvlms.
\newblock \emph{arXiv preprint arXiv:2406.11833}, 2024{\natexlab{e}}.

\bibitem[Lu et~al.(2024)Lu, Liu, Zhang, Wang, Dong, Liu, Sun, Ren, Li, Sun, et~al.]{lu2024deepseek}
Haoyu Lu, Wen Liu, Bo Zhang, Bingxuan Wang, Kai Dong, Bo Liu, Jingxiang Sun, Tongzheng Ren, Zhuoshu Li, Yaofeng Sun, et~al.
\newblock Deepseek-vl: towards real-world vision-language understanding.
\newblock \emph{arXiv preprint arXiv:2403.05525}, 2024.

\bibitem[Lu et~al.(2022)Lu, Mishra, Xia, Qiu, Chang, Zhu, Tafjord, Clark, and Kalyan]{lu2022learn}
Pan Lu, Swaroop Mishra, Tony Xia, Liang Qiu, Kai-Wei Chang, Song-Chun Zhu, Oyvind Tafjord, Peter Clark, and Ashwin Kalyan.
\newblock Learn to explain: Multimodal reasoning via thought chains for science question answering.
\newblock In \emph{The 36th Conference on Neural Information Processing Systems (NeurIPS)}, 2022.

\bibitem[Lu et~al.(2023)Lu, Bansal, Xia, Liu, Li, Hajishirzi, Cheng, Chang, Galley, and Gao]{lu2023mathvista}
Pan Lu, Hritik Bansal, Tony Xia, Jiacheng Liu, Chunyuan Li, Hannaneh Hajishirzi, Hao Cheng, Kai-Wei Chang, Michel Galley, and Jianfeng Gao.
\newblock Mathvista: Evaluating mathematical reasoning of foundation models in visual contexts.
\newblock \emph{arXiv preprint arXiv:2310.02255}, 2023.

\bibitem[Masry et~al.(2022)Masry, Long, Tan, Joty, and Hoque]{masry2022chartqa}
Ahmed Masry, Do~Xuan Long, Jia~Qing Tan, Shafiq Joty, and Enamul Hoque.
\newblock Chartqa: A benchmark for question answering about charts with visual and logical reasoning.
\newblock \emph{arXiv preprint arXiv:2203.10244}, 2022.

\bibitem[McCarthy and Jarvis(2010)]{mccarthy2010mtld}
Philip~M McCarthy and Scott Jarvis.
\newblock Mtld, vocd-d, and hd-d: A validation study of sophisticated approaches to lexical diversity assessment.
\newblock \emph{Behavior research methods}, 42\penalty0 (2):\penalty0 381--392, 2010.

\bibitem[Meta(2024)]{llama3.2vision}
Meta.
\newblock Llama 3.2 vision.
\newblock 2024.

\bibitem[{OpenAI}(2023)]{gpt4v}
{OpenAI}.
\newblock Gpt-4v(ision) technical work and authors, 2023.
\newblock \url{https://openai.com/contributions/gpt-4v}, Last accessed on 2024-02-13.

\bibitem[Pi et~al.(2024{\natexlab{a}})Pi, Zhang, Han, Zhang, Pan, and Zhang]{pi2024personalized}
Renjie Pi, Jianshu Zhang, Tianyang Han, Jipeng Zhang, Rui Pan, and Tong Zhang.
\newblock Personalized visual instruction tuning.
\newblock \emph{arXiv preprint arXiv:2410.07113}, 2024{\natexlab{a}}.

\bibitem[Pi et~al.(2024{\natexlab{b}})Pi, Zhang, Zhang, Pan, Chen, and Zhang]{pi2024image}
Renjie Pi, Jianshu Zhang, Jipeng Zhang, Rui Pan, Zhekai Chen, and Tong Zhang.
\newblock Image textualization: An automatic framework for creating accurate and detailed image descriptions.
\newblock \emph{arXiv preprint arXiv:2406.07502}, 2024{\natexlab{b}}.

\bibitem[Rahmanzadehgervi et~al.(2024)Rahmanzadehgervi, Bolton, Taesiri, and Nguyen]{rahmanzadehgervi2024vision}
Pooyan Rahmanzadehgervi, Logan Bolton, Mohammad~Reza Taesiri, and Anh~Totti Nguyen.
\newblock Vision language models are blind.
\newblock \emph{arXiv preprint arXiv:2407.06581}, 2024.

\bibitem[Shi et~al.(2024)Shi, Liu, Wang, Liao, Radhakrishnan, Huang, Yin, Sapra, Yacoob, Shi, et~al.]{shi2024eagle}
Min Shi, Fuxiao Liu, Shihao Wang, Shijia Liao, Subhashree Radhakrishnan, De-An Huang, Hongxu Yin, Karan Sapra, Yaser Yacoob, Humphrey Shi, et~al.
\newblock Eagle: Exploring the design space for multimodal llms with mixture of encoders.
\newblock \emph{arXiv preprint arXiv:2408.15998}, 2024.

\bibitem[Sirdeshmukh et~al.(2025)Sirdeshmukh, Deshpande, Mols, Jin, Cardona, Lee, Kritz, Primack, Yue, and Xing]{sirdeshmukh2025multichallenge}
Ved Sirdeshmukh, Kaustubh Deshpande, Johannes Mols, Lifeng Jin, Ed-Yeremai Cardona, Dean Lee, Jeremy Kritz, Willow Primack, Summer Yue, and Chen Xing.
\newblock Multichallenge: A realistic multi-turn conversation evaluation benchmark challenging to frontier llms.
\newblock \emph{arXiv preprint arXiv:2501.17399}, 2025.

\bibitem[Srinivasan et~al.(2021)Srinivasan, Raman, Chen, Bendersky, and Najork]{srinivasan2021wit}
Krishna Srinivasan, Karthik Raman, Jiecao Chen, Michael Bendersky, and Marc Najork.
\newblock Wit: Wikipedia-based image text dataset for multimodal multilingual machine learning.
\newblock In \emph{Proceedings of the 44th international ACM SIGIR conference on research and development in information retrieval}, pages 2443--2449, 2021.

\bibitem[Team et~al.(2023)Team, Anil, Borgeaud, Alayrac, Yu, Soricut, Schalkwyk, Dai, Hauth, Millican, et~al.]{team2023gemini}
Gemini Team, Rohan Anil, Sebastian Borgeaud, Jean-Baptiste Alayrac, Jiahui Yu, Radu Soricut, Johan Schalkwyk, Andrew~M Dai, Anja Hauth, Katie Millican, et~al.
\newblock Gemini: a family of highly capable multimodal models.
\newblock \emph{arXiv preprint arXiv:2312.11805}, 2023.

\bibitem[Tong et~al.(2024{\natexlab{a}})Tong, Brown, Wu, Woo, Middepogu, Akula, Yang, Yang, Iyer, Pan, et~al.]{tong2024cambrian}
Shengbang Tong, Ellis Brown, Penghao Wu, Sanghyun Woo, Manoj Middepogu, Sai~Charitha Akula, Jihan Yang, Shusheng Yang, Adithya Iyer, Xichen Pan, et~al.
\newblock Cambrian-1: A fully open, vision-centric exploration of multimodal llms.
\newblock \emph{arXiv preprint arXiv:2406.16860}, 2024{\natexlab{a}}.

\bibitem[Tong et~al.(2024{\natexlab{b}})Tong, Liu, Zhai, Ma, LeCun, and Xie]{tong2024eyes}
Shengbang Tong, Zhuang Liu, Yuexiang Zhai, Yi Ma, Yann LeCun, and Saining Xie.
\newblock Eyes wide shut? exploring the visual shortcomings of multimodal llms.
\newblock In \emph{Proceedings of the IEEE/CVF Conference on Computer Vision and Pattern Recognition}, pages 9568--9578, 2024{\natexlab{b}}.

\bibitem[Wang et~al.(2024{\natexlab{a}})Wang, Bai, Tan, Wang, Fan, Bai, Chen, Liu, Wang, Ge, Fan, Dang, Du, Ren, Men, Liu, Zhou, Zhou, and Lin]{Qwen2VL}
Peng Wang, Shuai Bai, Sinan Tan, Shijie Wang, Zhihao Fan, Jinze Bai, Keqin Chen, Xuejing Liu, Jialin Wang, Wenbin Ge, Yang Fan, Kai Dang, Mengfei Du, Xuancheng Ren, Rui Men, Dayiheng Liu, Chang Zhou, Jingren Zhou, and Junyang Lin.
\newblock Qwen2-vl: Enhancing vision-language model's perception of the world at any resolution.
\newblock \emph{arXiv preprint arXiv:2409.12191}, 2024{\natexlab{a}}.

\bibitem[Wang et~al.(2024{\natexlab{b}})Wang, Bai, Tan, Wang, Fan, Bai, Chen, Liu, Wang, Ge, et~al.]{wang2024qwen2}
Peng Wang, Shuai Bai, Sinan Tan, Shijie Wang, Zhihao Fan, Jinze Bai, Keqin Chen, Xuejing Liu, Jialin Wang, Wenbin Ge, et~al.
\newblock Qwen2-vl: Enhancing vision-language model's perception of the world at any resolution.
\newblock \emph{arXiv preprint arXiv:2409.12191}, 2024{\natexlab{b}}.

\bibitem[Wang et~al.(2023)Wang, Wang, Liu, Chen, Yuan, Peng, and Ji]{wang2023mint}
Xingyao Wang, Zihan Wang, Jiateng Liu, Yangyi Chen, Lifan Yuan, Hao Peng, and Heng Ji.
\newblock Mint: Evaluating llms in multi-turn interaction with tools and language feedback.
\newblock \emph{arXiv preprint arXiv:2309.10691}, 2023.

\bibitem[Wang et~al.(2024{\natexlab{c}})Wang, Xia, He, Chen, Liu, Zhu, Liang, Wu, Liu, Malladi, et~al.]{wang2024charxiv}
Zirui Wang, Mengzhou Xia, Luxi He, Howard Chen, Yitao Liu, Richard Zhu, Kaiqu Liang, Xindi Wu, Haotian Liu, Sadhika Malladi, et~al.
\newblock Charxiv: Charting gaps in realistic chart understanding in multimodal llms.
\newblock \emph{arXiv preprint arXiv:2406.18521}, 2024{\natexlab{c}}.

\bibitem[Wu et~al.(2023)Wu, Zhang, Zhang, Chen, Liao, Wang, Li, Sun, Yan, Zhai, et~al.]{wu2023q}
Haoning Wu, Zicheng Zhang, Erli Zhang, Chaofeng Chen, Liang Liao, Annan Wang, Chunyi Li, Wenxiu Sun, Qiong Yan, Guangtao Zhai, et~al.
\newblock Q-bench: A benchmark for general-purpose foundation models on low-level vision.
\newblock \emph{arXiv preprint arXiv:2309.14181}, 2023.

\bibitem[Xie et~al.(2024)Xie, Mao, Bai, Zhang, Wang, Lin, Gu, Chen, Yang, and Shou]{xie2024show}
Jinheng Xie, Weijia Mao, Zechen Bai, David~Junhao Zhang, Weihao Wang, Kevin~Qinghong Lin, Yuchao Gu, Zhijie Chen, Zhenheng Yang, and Mike~Zheng Shou.
\newblock Show-o: One single transformer to unify multimodal understanding and generation.
\newblock \emph{arXiv preprint arXiv:2408.12528}, 2024.

\bibitem[Xue et~al.(2024)Xue, Shu, Awadalla, Wang, Yan, Purushwalkam, Zhou, Prabhu, Dai, Ryoo, et~al.]{xue2024xgen}
Le Xue, Manli Shu, Anas Awadalla, Jun Wang, An Yan, Senthil Purushwalkam, Honglu Zhou, Viraj Prabhu, Yutong Dai, Michael~S Ryoo, et~al.
\newblock xgen-mm (blip-3): A family of open large multimodal models.
\newblock \emph{arXiv preprint arXiv:2408.08872}, 2024.

\bibitem[Yu et~al.(2023)Yu, Yang, Li, Wang, Lin, Liu, Wang, and Wang]{yu2023mm}
Weihao Yu, Zhengyuan Yang, Linjie Li, Jianfeng Wang, Kevin Lin, Zicheng Liu, Xinchao Wang, and Lijuan Wang.
\newblock Mm-vet: Evaluating large multimodal models for integrated capabilities.
\newblock \emph{arXiv preprint arXiv:2308.02490}, 2023.

\bibitem[Yu et~al.(2024)Yu, Yang, Ren, Li, Wang, Lin, Lin, Liu, Wang, and Wang]{yu2024mm}
Weihao Yu, Zhengyuan Yang, Linfeng Ren, Linjie Li, Jianfeng Wang, Kevin Lin, Chung-Ching Lin, Zicheng Liu, Lijuan Wang, and Xinchao Wang.
\newblock Mm-vet v2: A challenging benchmark to evaluate large multimodal models for integrated capabilities.
\newblock \emph{arXiv preprint arXiv:2408.00765}, 2024.

\bibitem[Yue et~al.(2024{\natexlab{a}})Yue, Ni, Zhang, Zheng, Liu, Zhang, Stevens, Jiang, Ren, Sun, et~al.]{yue2024mmmu}
Xiang Yue, Yuansheng Ni, Kai Zhang, Tianyu Zheng, Ruoqi Liu, Ge Zhang, Samuel Stevens, Dongfu Jiang, Weiming Ren, Yuxuan Sun, et~al.
\newblock Mmmu: A massive multi-discipline multimodal understanding and reasoning benchmark for expert agi.
\newblock In \emph{Proceedings of the IEEE/CVF Conference on Computer Vision and Pattern Recognition}, pages 9556--9567, 2024{\natexlab{a}}.

\bibitem[Yue et~al.(2024{\natexlab{b}})Yue, Zheng, Ni, Wang, Zhang, Tong, Sun, Yin, Yu, Zhang, et~al.]{yue2024mmmupro}
Xiang Yue, Tianyu Zheng, Yuansheng Ni, Yubo Wang, Kai Zhang, Shengbang Tong, Yuxuan Sun, Ming Yin, Botao Yu, Ge Zhang, et~al.
\newblock Mmmu-pro: A more robust multi-discipline multimodal understanding benchmark.
\newblock \emph{arXiv preprint arXiv:2409.02813}, 2024{\natexlab{b}}.

\bibitem[Zhang et~al.(2025)Zhang, Yao, Pi, Liang, and Fung]{zhang2025vlm2}
Jianshu Zhang, Dongyu Yao, Renjie Pi, Paul~Pu Liang, and Yi~R Fung.
\newblock Vlm2-bench: A closer look at how well vlms implicitly link explicit matching visual cues.
\newblock \emph{arXiv preprint arXiv:2502.12084}, 2025.

\bibitem[Zhang et~al.(2024{\natexlab{a}})Zhang, Jiang, Zhang, Lin, Guo, Qiu, Zhou, Lu, Chang, Gao, et~al.]{zhang2024mathverse}
Renrui Zhang, Dongzhi Jiang, Yichi Zhang, Haokun Lin, Ziyu Guo, Pengshuo Qiu, Aojun Zhou, Pan Lu, Kai-Wei Chang, Peng Gao, et~al.
\newblock Mathverse: Does your multi-modal llm truly see the diagrams in visual math problems?
\newblock \emph{arXiv preprint arXiv:2403.14624}, 2024{\natexlab{a}}.

\bibitem[Zhang et~al.(2024{\natexlab{b}})Zhang, Zhang, Tian, Fu, Zhang, Wu, Li, Wang, Wen, Zhang, et~al.]{zhang2024mme}
Yi-Fan Zhang, Huanyu Zhang, Haochen Tian, Chaoyou Fu, Shuangqing Zhang, Junfei Wu, Feng Li, Kun Wang, Qingsong Wen, Zhang Zhang, et~al.
\newblock Mme-realworld: Could your multimodal llm challenge high-resolution real-world scenarios that are difficult for humans?
\newblock \emph{arXiv preprint arXiv:2408.13257}, 2024{\natexlab{b}}.

\bibitem[Zheng et~al.(2023)Zheng, Chiang, Sheng, Zhuang, Wu, Zhuang, Lin, Li, Li, Xing, et~al.]{zheng2023judging}
Lianmin Zheng, Wei-Lin Chiang, Ying Sheng, Siyuan Zhuang, Zhanghao Wu, Yonghao Zhuang, Zi Lin, Zhuohan Li, Dacheng Li, Eric Xing, et~al.
\newblock Judging llm-as-a-judge with mt-bench and chatbot arena.
\newblock \emph{Advances in Neural Information Processing Systems}, 36:\penalty0 46595--46623, 2023.

\bibitem[Zheng et~al.(2024)Zheng, Huang, Xue, Wang, An, and Yan]{zheng2024agentstudio}
Longtao Zheng, Zhiyuan Huang, Zhenghai Xue, Xinrun Wang, Bo An, and Shuicheng Yan.
\newblock Agentstudio: A toolkit for building general virtual agents.
\newblock \emph{arXiv preprint arXiv:2403.17918}, 2024.

\end{thebibliography}
}

\clearpage
\appendix
\onecolumn

\section*{Appendix Contents}
\setcounter{tocdepth}{2}
\renewcommand{\contentsname}{Appendix Contents}
\startcontents[appendix]  
\printcontents[appendix]{}{1}{}

\clearpage

\section{Additional Related Works} \label{supp_sec:additional_related_work}

Recent advancements in LLVMs have predominantly adopted simplistic yet highly effective architectures, notably through the model-stitching concept. Numerous prior studies have introduced various design modifications to bridge the performance gap with closed-source LLVMs, such as GPT-4v~\citep{gpt4v}, GPT-4o~\cite{gpt4o}, Gemini-Pro~\cite{team2023gemini}, and Claude-3 family~\cite{claude3series2024}. These efforts include focusing intently on high-resolution processing~\citep{li2024mini,liu2024llava,shi2024eagle}, implementing locality-enhanced projectors~\citep{cha2024honeybee}, and incorporating knowledge embeddings~\citep{lee2024meteor}, layer traversal technique~\citep{lee2024trol} and leveraging a diverse array of vision encoders~\citep{lu2024deepseek,tong2024eyes} have also been explored. Additionally, integrating external, task-specific computer vision modules~\citep{lee2024collavo,lee2024moai,jiao2024enhancing,lai2024lisa} and incorporating different modalities — including video and audio~\citep{wang2024qwen2,li2024llavaone,sharegpt4o,xie2024show} — have expanded the models' capabilities. Moreover, enabling the handling of interleaved input formats~\citep{li2024llava,xue2024xgen} has further broadened the versatility of these models.

\section{Additional Details of \datasetName} \label{supp_sec:additional_benchmark}

{\renewcommand{\arraystretch}{1.35}
\begin{table*}[th!]
\centering
\resizebox{1.0\linewidth}{!}{
\begin{tabular}{l|c|l|l}
\toprule
\textbf{Datasets}     & \textbf{\# Images} & \textbf{Public Links}                                                                            & \textbf{Licenses}            \\ \midrule
MMDU~\cite{liu2024mmdu}         & 375                           & \url{ https://huggingface.co/datasets/laolao77/MMDU}             & CC-BY-NC-4.0       \\
MegaBench~\cite{chen2024mega}    & 12091                         & \url{ https://huggingface.co/datasets/TIGER-Lab/MEGA-Bench}      & Apache License 2.0 \\
GroundUI-1K~\cite{zheng2024agentstudio}  & 1000                          & \url{ https://huggingface.co/datasets/agent-studio/GroundUI-1K}  & MIT                \\
MMMU~\cite{yue2024mmmu}         & 12141                         & \url{ https://huggingface.co/datasets/MMMU/MMMU}                 & Apache License 2.0 \\
MMMU-Pro~\cite{yue2024mmmupro}     & 2015                          & \url{ https://huggingface.co/datasets/MMMU/MMMU\_Pro}            & Apache License 2.0 \\
NaturalBench~\cite{li2024naturalbench} & 3800                          & \url{ https://huggingface.co/datasets/BaiqiL/NaturalBench}       & Apache License 2.0 \\
VisIT-Bench~\cite{bitton2023visit}  & 574                           & \url{ https://huggingface.co/datasets/mlfoundations/VisIT-Bench} & CC-BY-4.0          \\
MathVista~\cite{lu2023mathvista}    & 6141                          & \url{ https://huggingface.co/datasets/AI4Math/MathVista}         & CC-BY-SA-4.0       \\
MM-Vet~\cite{yu2023mm}       & 218                           & \url{ https://huggingface.co/datasets/whyu/mm-vet}               & CC-BY-NC-4.0       \\
MM-Vet v2~\cite{yu2024mm}    & 704                           & \url{ https://huggingface.co/datasets/whyu/mm-vet-v2}            & CC-BY-NC-4.0       \\
CharXiv~\cite{wang2024charxiv}      & 2323                          & \url{ https://huggingface.co/datasets/princeton-nlp/CharXiv}     & CC-BY-SA-4.0       \\
MMBench (Eng)~\cite{liu2023mmbench}      & 6718                          & \url{ https://huggingface.co/datasets/lmms-lab/MMBench\_EN}      & Apache License 2.0 \\ \midrule
\textbf{All Images} & \textbf{48100} & - & - \\
\bottomrule

\end{tabular}}
\caption{Detailed information on the 16 source datasets collected for \datasetName.}
\label{supp_tab:source_dataset_statistics}
\end{table*}
}

In Step 1, we collect 55,500 images from 16 existing evaluation benchmarks. The detailed statistics, including the number of images, public links, and license information, are provided in Table~\ref{supp_tab:source_dataset_statistics}. 

\paragraph{License.} All evaluation benchmarks permit non-commercial use, allowing us to preprocess and redistribute the data. However, certain datasets, such as MathVista and CharXiv, are licensed under \texttt{CC-BY-SA-4.0}. This requires us to comply with the same licensing terms when publicly deploying \datasetName.

\clearpage
\section{Further Analysis of \datasetName} \label{supp_sec:mildconv_analysis}

\subsection{Interaction Goal}

\paragraph{Verification (87)}: Data Validation, Data Consistency, Statistical Significance, Financial Verification, Graph Intersection Points, Cost Accounting Verification, Strategy Validation, Equation and Calculation Verification, Answer Verification, Image Details Verification, Term Definition Matching, Mathematical Theorem and Calculation Verification, Graph Interpretations, Verification of Calculations and Materials, Anatomical Details Verification, Molecular Structure and Properties, Design Verification, Mechanical Dynamics Calculations, Image Verification, Graph Interpretation, Sequence Evaluation, Dimension Verification, Financial Viability, Biopsy Image Analysis, Geometric Problem Verification, Feedback Control System Design, Data Interpretation Verification, Data Accuracy Verification, Equation Verification, Data Interpretation, Structural Calculations, Feedback, Musical Score Interpretation, Location and Meaning, Financial Analysis, Code Verification, Mathematical Properties, Solution Verification, Data Correlation, Graph Theory, Model Interpretation Verification, Interpretation, Alignment with Theoretical Models, Interpretation Verification, Geometric Properties, Molecular Structure Verification, Optimization, Geometry, Compatibility Verification, Geometric Verification, Data Accuracy, Audio Issue Resolution, Verifying Distribution Shape, Logic Verification, Calculation Verification, Data Analysis, Theoretical Implications, Identification, Music Theory, Theoretical Consistency, Molecular Structure, Mathematical Method, Structural Analysis Calculations, Safety and Performance Evaluation, Circuit Calculation Verification, Circuit Verification, Function Identification, Formula Verification, Algorithm Verification, Molecular Structure Confirmation, Feedback Collection, Data Verification, Financial Projections, Confirmation, Enhancement, Mathematical Verification, Feedback Assessment, Parameter Suitability, Diagnosis, Solution Checking, Signal Analysis Verification, Cost Verification, Interpretation Checking, Design Feedback, Thermodynamic Properties, Financial Statement Compliance, Accuracy Verification

\paragraph{Analysis (123)}: Problem Solving, Chess Endgame Strategy, Weather Patterns, Model Convergence in Option Pricing, Mechanical Stress Distribution, Stress Analysis, Art Analysis, Pattern Recognition, Data Comparison, Trend Analysis, Pattern Detection, Impact of Sparsity on Data Recovery, Disease Outbreak Factors, Risk-Return Trade-offs, Schedule Optimization, Connectivity Analysis, Graph Data Interpretation, Statistical Methods Comparison, Travel Data Analysis, Impact Assessment, Technique Evaluation, Performance Metrics, Performance Comparison, Classification, Music Analysis, Health Disparities, Investment Strategy Analysis, Cost and Time Implication Analysis, Cost Optimization, Model Interpretation, Economic Analysis, Chess Endgame Analysis, Feedback Evaluation, Data Interpretation, Probability Distribution Models, Performance Interpretation, Performance Analysis, Bias Detection, Design Elements Assessment, Financial Analysis, Game Evaluation, Data Pattern Interpretation, Architectural Analysis, Economic Impact, Model Performance Evaluation, Geometric Pattern Analysis, Mechanical Forces, Diagnosing Plant Issues, Strategic Planning, Artistic Elements and Symbolism in Manga, Design Evaluation, Material Evaluation, Price Comparison, Trade-offs Assessment, Measurement Suitability, Causation Analysis, Interpretation, Geometric Relationships Analysis, Traffic and Urban Design, Investment Options, System Behavior Analysis, Behavioral Analysis and Health Assessment, Medical Condition Assessment, Project Management Insights, Symbolism Identification, Equity Changes, Force Distribution Study, Property Interpretation, Artistic Analysis, Strategy Evaluation, Risk-Return Trade-off, Ecosystem Health Assessment, Demographic Analysis, Financial Health Analysis, Data Analysis, Decision-making Analysis, Identification, Correlation Analysis, Risk and Return, Musical Analysis, Suitability Assessment, Design Elements, Trend Identification, Curve Identification, Strategic Analysis, Optimization Analysis, Statistical Analysis, Memory Management, Market Research, Function Identification, Spectral Analysis, Electricity Usage Analysis, Comparative Analysis, Performance and Robustness Evaluation, Enhancing Engagement Strategies, Summarization, Nuclear Reaction Analysis, Comparative Evaluation, Data Trends, Chess Strategy, Quantum Mechanics, Function Behavior Analysis, Financial Market Analysis, Cause Identification, System Stability, Research Analysis, Diagnosis, Empirical vs Theoretical Comparison, Investment Strategy Performance, Damage Assessment, Sales Trend Analysis, Medical Analysis, Infection Spread Dynamics, Interpretation of Experimental Data, Pattern Analysis, Sustainability, Visual Techniques, Comparison, Performance Metrics Interpretation, Data Visualization, Feedback on Composition, Correlation Study, Stock Market Trends

\paragraph{Exploration (30)}: Instructional Strategies, Idea Generation, Cultural and Historical Exploration, Geometric Properties, Design Elements, Information Gathering, Cultural and Historical Insights, Music Analysis, Ingredient Identification, Music Discovery, Weather and Photography, Geometric Structure Identification, Applications of Geometric Art in Architecture, Design Improvements, Algorithm Development, Species Identification, Model Recommendations, Design and Aesthetics, Comparison and Analysis of Options, Puzzle Solving Strategies, Interpretation of Visual Elements, Feature Discovery, Species Recognition, Assessment of Urban Environmental Features, Design Variations, Art and Design Inspiration, Chemical Compound Identification, Information Discovery, Creative Applications, Coordinate Conversion

\paragraph{Optimization (48)}: Process Improvement, Readability Enhancement, Algorithm Efficiency, Strategic Improvement, Project Management, Scheduling, Material Selection, Code Quality, Design Efficiency, Pathfinding, Translation Model Performance Enhancement, Improving Visual Appearance, Gameplay Strategy, Code Efficiency, Neural Network Configuration, Team Building, Cost Optimization, Image Enhancement, Inventory Management, Visualization Clarity, Decision Making, Route Efficiency, Rendering Techniques Improvement, Strategy Development, Code Optimization, Process Optimization, Eco-friendliness, Investment Portfolio, Clarity Enhancement, Pricing Strategy Optimization, Integration of auditory and visual elements, Schedule Optimization, Resource Allocation, Enhancing Real Estate Listings, Visual Design, System Efficiency Enhancement, Strategic Planning, Code Efficiency and Readability, Price Comparison, Resource Management, Workflow Efficiency, Organization, Improve Data Visualization, Script Reliability, Data Visualization, Robot Movement Strategy, Efficiency Enhancement, Vibration Isolation Methods

\paragraph{Calculation (32)}: Financial Calculation, Problem Solving, Arithmetic Problem, Project Management, Geometry Problem Solving, Geometric Calculation, Investment Evaluation, Geometry, Financial Computation, Structural Analysis, Force Calculation, Finance, Cost Analysis, Decision Making, Investment Returns, Weighted Average Calculation, Pressure Difference Calculation, Geometry Calculation, Financial Analysis, Investment Performance Analysis, Equilibrium Calculation, Trajectory Calculation, Load Analysis, Financial Modeling, Statistical Metric Calculation, Solving Equations, Interpolation, Structural Engineering, Cost Estimation, Stress Intensity Factors, Trigonometry, Geometrical Calculation

\paragraph{Understanding (86)}: Concept Clarification, Mechanics, Comprehension of Data Visualizations, Technical Concepts, Enrollment Disparities, Chemical Properties and Applications, Scientific Concepts, Model Performance, Implications of OOD detection in medical predictions, System Settings, Electrical Circuit Analysis, AI Capabilities, Mathematical Relationships, Material Impact, Comprehension of Concepts, Comprehension of Biological Systems, Comprehension of Graph Data, Thermodynamics Concepts, Solution Process, Educational Trends, Learning how to create geometric patterns, Conceptual Differences, Graph Interpretation, Deciphering, Structural Mechanics, Effects of Data Augmentation, Code Logic and Concepts, Economic Concepts, Behavior and Applications, Function Behavior, Seismic Data Interpretation, Simplification, Advanced Mathematics Problems, Understanding Electrostatic Potential, Data Interpretation, Performance Analysis, System Dynamics, Algorithm Logic Understanding, Comprehension of Construction Techniques, Ecological Roles, Explaining Properties and Applications, Comprehension of Theoretical Concepts, Graph Theory, Structural Engineering, Application of Frameworks, Automotive Dynamics, Historical Context, Geometric Properties, Historical and Cultural Analysis, Concept Explanation, Function and Importance of DVD Layers, Research Methodologies, Functionality Explanation, Plant Growth and Reproduction, Technology Applications, Symbolism Analysis, Sales Insights, Understanding Concurrency Issues, Music Theory Comprehension, Clarification, Historical and Architectural Significance, Geometric Concepts, Biology Concepts, Music Theory, Symbolism in Art, Application of Technology, Color Interaction and Emotional Impact, Circuit Configuration and Components, Cultural Significance, Conceptual Clarification, Conceptual Comprehension, Comparative Analysis, Comprehension of Map Projections, Comprehension of Tree Traversals, Code Structure and Functionality, Chemical Reactions, Behavior Analysis, Reinforcement Learning Policies, Conceptual Understanding, Terminology, Relationship Analysis, Clarification of Processes, Leadership Pyramid Levels, Algorithmic Optimization, Color Theory, Comprehension of Dynamics

\paragraph{Research (72)}: Data Retrieval, Historical and Route Information Retrieval, Teaching Techniques, Design Insights Gathering, Recipe Acquisition, Historical Information Gathering, Gathering Resources, Information Gathering, Product Identification and Supplier Sourcing, Historical and Architectural Research, Hotel Exploration, Historical Significance, Gathering Information, Historical and Architectural Information, Educational Resources, Economic Trends Analysis, Art Analysis, Historical and Technical Information, Cultural Studies, Information Retrieval, Biological Research, Trend Analysis, Data Interpretation, Species Identification, Art History, Architectural Patterns, Travel Logistics, Cultural and Historical Analysis, Trends Analysis, Historical Insights, Product Information Gathering, Safety Regulations and Design Practices, Ecological Study, Literature Review, Sports Analysis, Identification and Knowledge Acquisition, Art, Historical and Mythological Context, Ecological Information Gathering, Cultural and Historical Research, Film Analysis, Historical and Architectural Details, Gathering Expert Tips, Product Features and Benefits, Identification, Historical Research, Art Style Similarity, Symbiosis in Marine Biology, Design Origins, Historical Analysis, Horticulture, Mentorship and Collaboration Opportunities, Historical Events, Sustainable Design Insights, Historical and Architectural Analysis, Disease Identification and Management, Geological Study, Art Techniques, Historical Economic Impact, Comparison, Art and Artist Analysis, Art Historical Contextualization, Geology, Agricultural Knowledge Gathering, Historical Context, Historical Design Analysis, Cultural and Historical Facts Gathering, Ecology, Historical and Cultural Significance, Collaboration, Music History, Architectural Insights

\paragraph{Creation (43)}: Explain, Educational Activity Design, Interior Design, Creative Cooking, Design Brainstorming, Design Enhancement, Lyric Writing, Chart Creation, Infographic Design, Lesson Plan Design, Word Association, Design Inspiration, Content Generation, Activity Design, Schedule Design, Design Integration, Developing Photography Exhibit Concepts, Educational Content Development, Interactive Educational Activities Design, Guide Development, Design Generation, Question Development, Content Creation, Educational Material Development, Exercise Development, Data Compilation, Digital Transcription, Discussion Generation, Humor Enhancement, Culinary Innovation, Culinary Creation, Material Enhancement, Recipe Development, Creative Ideation, Cocktail Ingredient Arrangement, Data Visualization, Demonstration Development, Game Development, Data Visualization Enhancement, Engagement, Culinary Innovating, Visualization, Artistic Inspiration and Design

\section{More examples on \datasetName} \label{supp_sec:mildconv_examples}

\begin{figure}[t]
    \centering
    \includegraphics[width=0.9\linewidth]{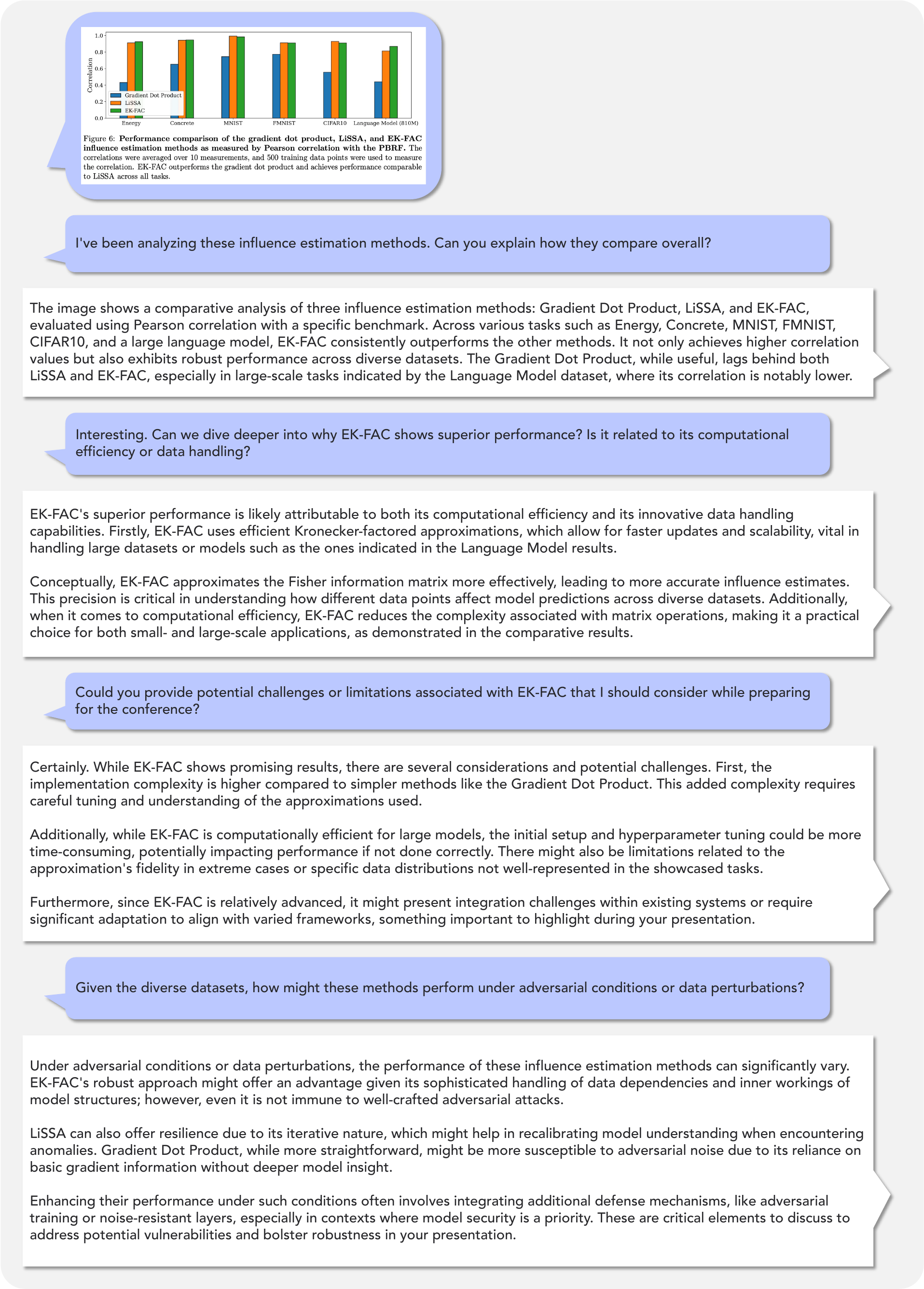}
    \caption{An example of \datasetName.}
    \label{supp_fig:case1}
\end{figure}

\begin{figure}[t]
    \centering
    \includegraphics[width=0.9\linewidth]{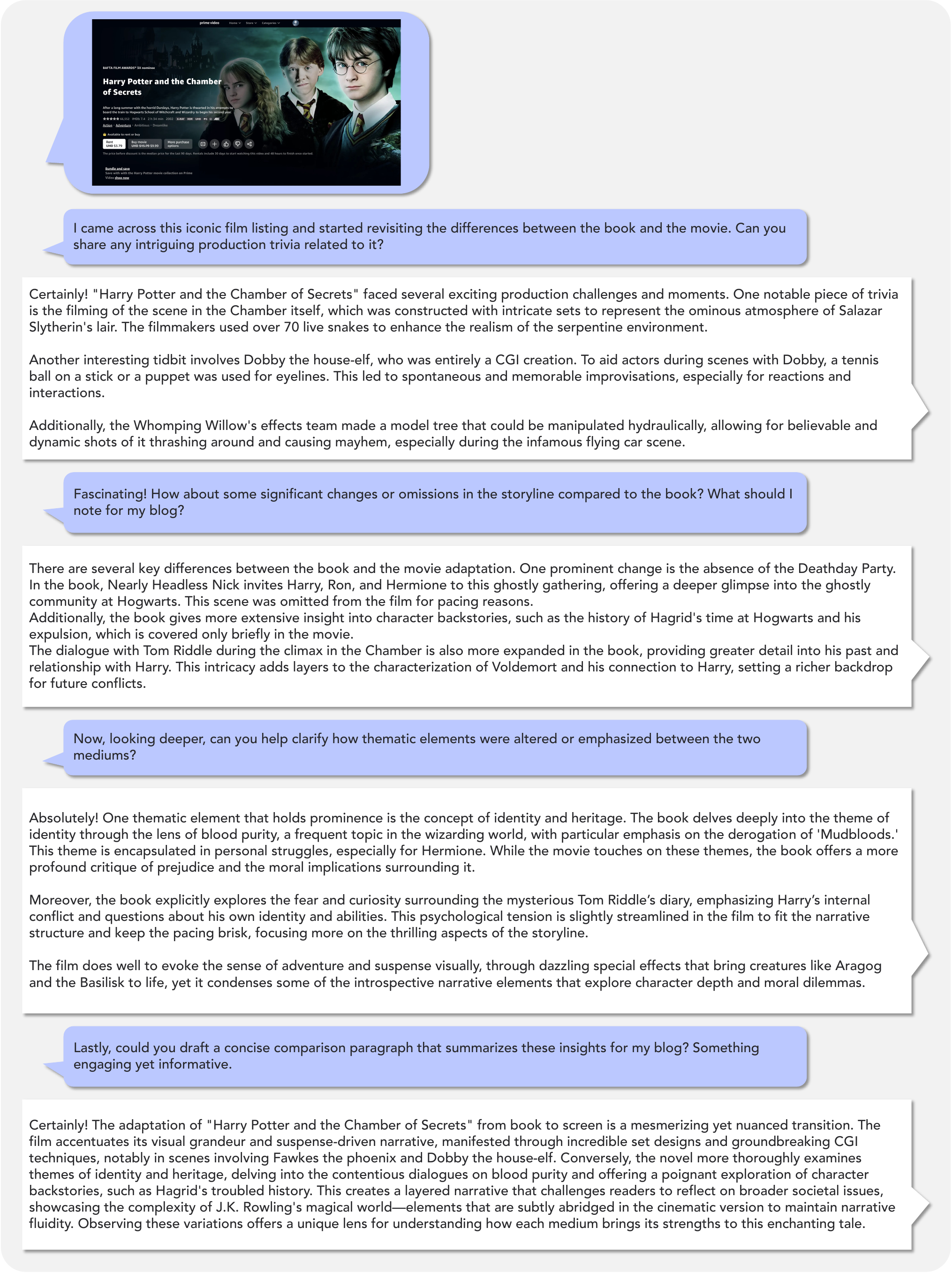}
    \caption{An example of \datasetName.}
    \label{supp_fig:case2}
\end{figure}

\begin{figure}[t]
    \centering
    \includegraphics[width=0.9\linewidth]{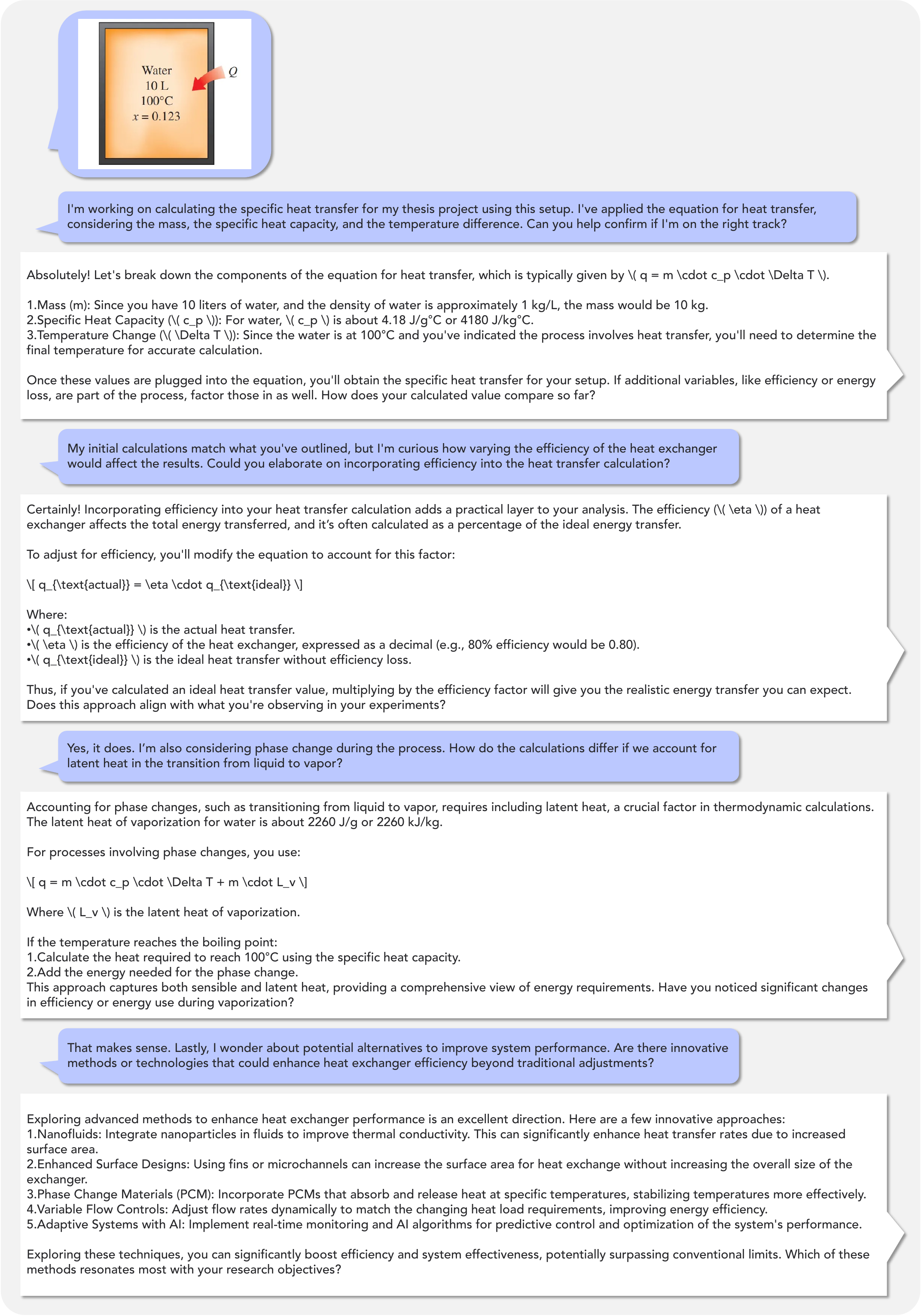}
    \caption{An example of \datasetName.}
    \label{supp_fig:case3}
\end{figure}

\begin{figure}[t]
    \centering
    \includegraphics[width=0.9\linewidth]{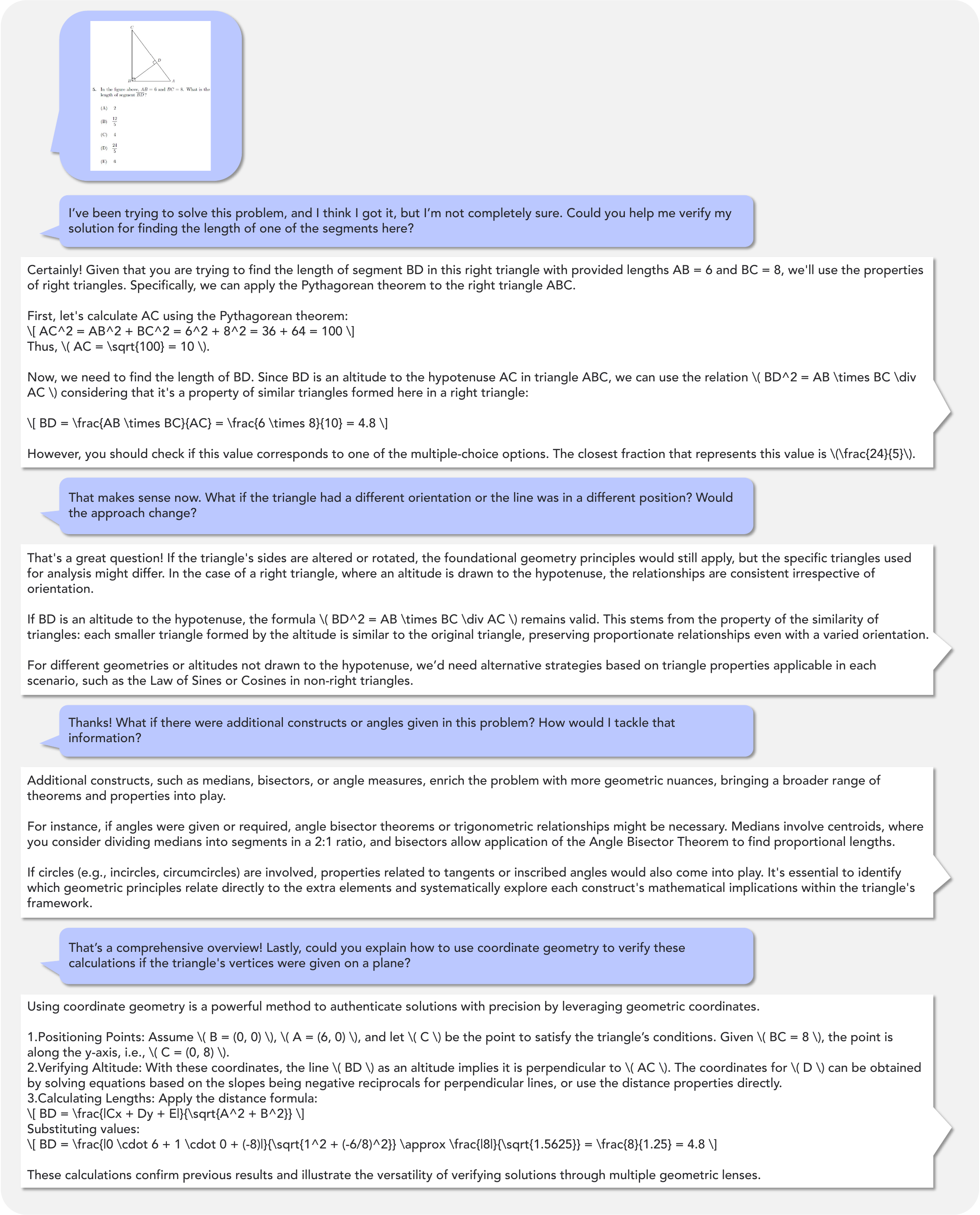}
    \caption{An example of \datasetName.}
    \label{supp_fig:case4}
\end{figure}

We present more examples on \datasetName in Figure~\ref{supp_fig:case1}, Figure~\ref{supp_fig:case2}, Figure~\ref{supp_fig:case3}, and Figure~\ref{supp_fig:case4}.

\clearpage
\section{Prompt Templates used for \datasetName} \label{supp_sec:mildconv_prompt}

\begin{prompt}{Prompt Template for Scoring Image Quality}
    You are given an image. Your task is to evaluate the image quality, which is described as follows: \\
    \\
    \lbrack Image Quality Evaluation\rbrack \\
    Rate the given image on a scale from 1 to 5 based on its quality, considering image clarity, resolution, and likelihood of occurrence in real-world conversations: \\
    - Score 1 (Very Low Quality): The image lacks clarity, and objects or content appear blurry with minimal detail, making it unlikely to appear in real conversations. Additionally, this score applies if the image includes a specific brand logo or icon or if the image itself has been rotated. \\
    - Score 2 (Low Quality): The image is not sharp, with only some objects or elements visible, and is rarely used in real conversations. \\
    - Score 3 (Moderate Quality): The image is moderately clear in parts, with sufficient visibility of objects or content to be recognized in specific situations or by certain individuals in conversation. \\
    - Score 4 (High Quality): The image is clear, with most objects or elements recognizable, and is likely to appear in real conversations. \\
    - Score 5 (Very High Quality): The image is extremely clear, with even small objects or details highly visible, making it highly relevant for frequent use in real conversations. \\
    \\
    Please generate your answer by strictly following the guidelines below: \\
    \\
    \lbrack Guidelines\rbrack \\
    - The answer should be formatted as a Python dictionary containing the following key: ``image\_quality\_score''. \\
    - The ``image\_quality\_score'' should indicate the quality of the given image, range from 1 to 5. \\
    \\
    \lbrack Answer\rbrack \\
\end{prompt}

\begin{prompt}{Prompt Template for Image Category}
    You are given an image. Your task is to assign a subject/category to the image, which is described as follows: \\
    \\
    \lbrack Image Category Classification\rbrack \\
    Classify the subject/category of the given image based on its content. Choose from the following categories, along with the corresponding descriptions: \\
    - Vehicles and Transportation: Includes all types of transportation, such as cars, bikes, trains, planes, and ships. \\
    - Food and Cuisine: Covers cuisine styles, dishes, ingredients, and cooking activities. \\
    - People and Lifestyle: Encompasses everyday life, cultural lifestyles, and social activities involving people. \\
    - Sports and Recreation: Includes all sports, games, and leisure activities. \\
    - Animals and Wildlife: Covers both domestic animals and wildlife, with subcategories like birds, sea creatures, etc. \\
    - Objects, Clothing, and Accessories: Encompasses fashion, personal items, and any identifiable object. \\
    - Brands and Products: Relates to consumer goods, brand logos, and popular products. \\
    - Architecture and Landmarks: Includes buildings, bridges, monuments, and notable geographic features. \\
    - Tradition and History: Covers historical sites, artifacts, and traditional practices. \\
    - Fine Art and Illustrations: Includes classical paintings, digital art, and creative illustrations. \\
    - Celebrities and Public Figures: Specific to well-known individuals in popular culture or politics. \\
    - Science and Technology: Includes scientific equipment, laboratories, and research environments. \\
    - Chemistry and Lab Equipment: Covers chemicals, laboratory settings, and molecular structures. \\
    - Mathematics and Symbols: Encompasses mathematical diagrams, equations, and geometric shapes. \\
    - Nature and Landscapes: Includes scenery, forests, mountains, rivers, and natural phenomena. \\
    - Healthcare and Medicine: Covers medical devices, doctors, hospitals, and health-related content. \\
    - Programming and Coding: Depicts code, software interfaces, and development environments. \\
    - Web Design and User Interfaces: Encompasses UI/UX layouts, website mockups, and interface elements. \\
    - Mobile and Smart Devices: Includes smartphones, tablets, and other handheld devices. \\
    - Weather and Climate: Covers weather events, climate patterns, and related data visualization. \\
    - Festivals and Events: Encompasses celebrations, public gatherings, and cultural events. \\
    - Education and Learning: Covers classrooms, books, e-learning visuals, and study materials. \\
    - Entertainment and Media: Includes visuals from movies, shows, music, and gaming. \\
    - Interior Design and Furniture: Relates to home decor, furniture styles, and interior layouts. \\
    - Natural Disasters and Environmental Hazards: Includes earthquakes, floods, pollution, and environmental issues. \\
    - Abstract and Conceptual Art: Covers non-representational or symbolic images. \\
    - Business and Finance: Depicts financial graphs, business meetings, currency, and economy-related visuals. \\
    - Law and Government: Includes courts, governmental symbols, legal documentation, and policy. \\
    - Diagrams and Schematics: Covers flowcharts, technical diagrams, mind maps, and other schematic representations. \\
    - Charts and Graphs: Encompasses data visualizations like bar charts, pie charts, line graphs, and other statistical representations. \\
    - Documents and Paperwork: Covers text-heavy documents, official forms, written notes, and reports. \\
    - Receipts and Invoices: Includes financial documents, bills, purchase receipts, and transaction records. \\
    - Others \\
    \\
    Please generate your answer by strictly following the guidelines below: \\
    \\
    \lbrack Guidelines\rbrack \\
    - The answer should be formatted as a Python dictionary containing the following key: ``image\_category'', and ``image\_sub\_category''. \\
    - The ``image\_category'' should contain the most appropriate classified category of the given image. \\
    - The ``image\_sub\_category'' should specify a more detailed sub-category within the selected ``image\_category'' to provide a finer level of classification. \\
    - If you select ``Others'', then please generate a new image category. \\
    \\
    \lbrack Answer\rbrack \\
\end{prompt}

\begin{prompt}{Prompt Template for Personal Background Generation}
    You will be provided with an image. \\
    \\
    Your task is to: \\
    \\
    1. Fictional Character Creation \\
    - You should create a fictional character who plausibly engages in a conversational interface with an AI assistant by providing a given image. \\ 
    - The fictional character should initiate a realistic and contextually appropriate conversation based on the image. \\
    - The fictional character should be represented in 2-3 sentences (NOT a structured format), covering aspects like name, age, gender, occupation, preference, hobby, like/dislike, or background knowledge (e.g., basic, intermediate, expert). \\
    - If there is a real human in the given image, please do NOT generate the fictional character corresponding to the real human in the given image. You MUST generate a new fictional character who will share the given image with the AI assistant in the conversational interface. \\
    \\
    2. Scenario Context Creation \\
    - You should create a plausible and detailed scenario in which the fictional character interacts with the AI assistant about the given image through a conversational interface. \\
    - The scenario context should realistically unfold in a multi-turn conversation between the fictional character and the AI assistant. \\
    - The scenario should be practical, realistic, and engaging, reflecting real-world situations. \\
    \\
    3. Goal Creation \\
    - You should create a specific and realistic goal that the fictional character has when engaging in a conversation with the AI assistant through a conversational interface by providing a given image, in alignment with the generated scenario context. \\
    - The goal should be relevant to the scenario context and represented as a concise phrase. For example, the assistant might search for information, retrieve data, solve a math or coding problem, etc. \\
    \\
    Please brainstorm the most appropriate, realistic, and highly plausible fictional character, scenario context, and goal that could naturally occur in a real-world conversation based on the given image by strictly following the JSON format below: \\
    - The answer should be formatted as a JSON-formatted Python dictionary containing the following keys: ``Character Description'', ``Scenario'', and ``Goal''. \\
    \\
    Answer:
\end{prompt}

\begin{prompt}{Prompt Template for Multi-Turn Conversation Generation}
    You will be provided with an image. \\
    \\
    Your task is to brainstorm a creative, realistic and practical multi-turn conversation between a fictional character (denoted as \lbrack Fictional Character\rbrack) and the AI assistant, based on a given scenario context (denoted as \lbrack Scenario Context\rbrack) and the fictional character’s goal (denoted as \lbrack Goal\rbrack). In the conversation, the fictional character interacts with the AI assistant through a conversational interface by providing an image. \\
    \\
    The fictional character, scenario context, and goal are presented as follows: \\
    \\
    \lbrack Fictional Character\rbrack \\
    \textcolor{blue}{\texttt{\{character\}}} \\
    \\
    \lbrack Scenario Context\rbrack \\
    \textcolor{blue}{\texttt{\{scenario\}}} \\
    \\
    \lbrack Goal\rbrack \\
    \textcolor{blue}{\texttt{\{goal\}}} \\
    \\
    Please brainstorm a practical and realistic multi-turn conversation based on the given fictional character, scenario context, goal, and the provided image by strictly following the guidelines below: \\
    \\
    \lbrack Guidelines\rbrack \\
    - The generated multi-turn conversation should be presented in plain text, where the fictional character's utterances MUST begin with ``USER:'', and the AI assistant's utterances MUST begin with ``ASSISTANT:''. \\
    - As the conversation progresses, the fictional character should ask progressively more challenging, creative, complex requests, such as follow-up questions, knowledge acquisition, refinement, style/content rephrasing, advanced reasoning expansion, etc. \\
    - The fictional character's utterances should be creative, reflect diverse linguistic styles, and include questions that require correct answers to difficult problems. \\
    - The AI assistant’s utterances MUST be very long detailed, specific, expert, helpful, and informative in addressing the fictional character’s requests. \\
    - The AI assistant's utterances should demonstrate advanced cognitive reasoning and do NOT include the fictional character's name. \\
    - If the AI assistant’s utterances involve code, mathematical equations, charts, tables, graphs, scientific concepts, API functions, planning, or any higher-order knowledge, they MUST be highly informative, expert, specific, detailed, and factual. They should also include clear and comprehensive explanations alongside the technical details (e.g., code, mathematical theorems, API function calls, etc.). \\
    - Each turn consists of two utterances: one from the fictional character and one from the AI assistant. \\
    - The conversation should not end with a closing remark like ``See you next time'' or similar. \\
    - The conversation must consist of four turns (eight utterances), with each turn involving one utterance from the fictional character and one from the AI assistant. \\
    \\
    Note: \\
    - NEVER generate utterances from characters in cases where the AI assistant can resolve them precisely without viewing the given image. \\
    - ALWAYS refer to the given image using pronouns (e.g., ``it,'' ``them'') in any fictional character's utterances. \\
    - NEVER explicitly mention the key entity or information depicted in the image in any fictional character's utterances. \\
    \\
    \lbrack Generated Multi-Turn Conversation\rbrack 
\end{prompt}

\begin{prompt}{Prompt Template for Checklist Generation}
    You will be provided with an image and the user's utterances from a multi-turn conversation. The conversation takes place in a conversational interface where the user and the AI assistant interact while referring to the given image. \\
    \\
    Your task is to create an instance-specific evaluation checklist that will be used to assess the quality of the AI assistant’s response to the user’s query within the multi-turn conversation. In other words, you need to create the evaluation criteria (in a question format) that the AI assistant’s response must meet to be considered the optimal response for the given user query, based on the given image. \\
    \\
    The checklist should include multiple questions, each satisfying the following conditions: \\
    - Each question must be answerable with ``Yes'' or ``No.'' \\
    - Each question must be relevant to one specific evaluation criteria from the provided criteria collection. \\
    - Each question should minimize subjectivity in the rater’s judgment. \\
    - Questions should be formulated so that a ``Yes'' answer is a positive evaluation. \\
    - Each question must evaluate aspects related to the conditions that need to be met—by referring to the given image and each turn of the user's query—to produce the optimal response, and these conditions must be strictly satisfied. \\
    \\
    \#\#\# Evaluation Criteria Collection \\
    Each evaluation criteria represents a main capability to evaluate (alongside the definition): \\
    - Engagingness: Measures the model’s ability to sustain an engaging and interactive conversation by assessing flow, immersion, interactivity, and emotional connection with the user. \\
    - Tone \& Style Appropriateness: Evaluates whether the response maintains an appropriate, positive, polite, and respectful tone. \\
    - Contextual Understanding: Assesses how well the model understands the previous conversational context, including anaphora resolution and maintaining consistency across multiple turns. \\
    - Memory: Evaluates whether the model accurately remembers and incorporates earlier dialogue details while tracking and retaining long-term contextual dependencies. \\
    - Proactiveness: Determines whether the model proactively identifies and fulfills user needs by making helpful suggestions and guiding the conversation. \\
    - Clarity \& Logical Structure: Evaluates the clarity, coherence, and logical organization of responses, ensuring they are easy to understand.
    - Coherence: Ensures responses maintain consistency across turns and that dialogue progresses naturally without contradictions. \\
    - Knowledge Understanding: Determines whether the model demonstrates expertise and incorporates relevant domain knowledge with depth and accuracy. \\
    - Factual Correctness: Ensures responses contain accurate, verifiable facts and do not generate misinformation. \\
    - Specificity \& Informativeness: Evaluates how specific and detailed the response is, measuring the amount of useful, non-generic information provided. \\
    - Cognitive Reasoning: Assesses the model’s ability to reason logically and make commonsense inferences, including problem-solving and inference-making. \\
    - Creativity: Evaluates the model’s ability to generate creative and original content in storytelling, idea generation, coding, and visual responses. \\
    - Problem-Solving Capability: Measures how effectively the model breaks down and addresses complex user queries, including step-by-step explanations. \\
    - Helpfulness: Determines whether the response directly addresses user needs and provides actionable suggestions. \\
    - Instruction Following: Assesses the model’s ability to interpret and adhere to explicit user instructions without deviation. \\
    - Harmlessness \& Ethical Awareness: Ensures responses avoid harmful, offensive, or biased content while adhering to ethical considerations. \\
    - Perceptual Understanding: Evaluates how well the model interprets the given image and integrates it into coherent responses. \\
    - Fluency \& Grammatical Accuracy: Assesses the linguistic correctness of responses, including grammar, spelling, sentence completeness, and verb tense consistency. \\
    - Adaptability: Measures how well the model adapts responses based on user background, preferences, and conversational history. \\
    - Multimodal Consistency: Ensures the model maintains semantic alignment between different modalities (e.g., text and images) and that textual responses accurately reflect visual inputs. \\
    \\
    \#\#\# Multi-Turn Conversation \\
    Below is the multi-turn conversation (the AI assistant's responses are regarded as the optimal answers to the user queries). The conversation always starts at turn number 1.  \\
    \\
    \textcolor{blue}{\texttt{\{conversation\}}} \\
    \\
    ---\\
    \\
    Please brainstorm comprehensive, creative, and practical evaluation checklists with as many relevant questions as possible for each user query (one checklist per user query). \\
    \\
    You MUST strictly follow the guidelines below: \\
    \\
    \lbrack Guidelines\rbrack \\
    - The output should be formatted as a JSON-formatted Python list.\\
    - Each entry in the list should be a Python dictionary containing the following keys: ``utterance\_id'' and ``checklist''.\\
    - The ``utterance\_id'' should indicate the utterance index of the user's utterance in the conversation, with the utterance index always starting at 1.\\
    - Each entry in the ``checklist'' should be a Python dictionary containing the following keys: ``question'', ``main\_criteria'', and ``sub\_criteria''.\\
    - The ``question'' should indicate a very detailed and specific question that should be answerable with the positive answer of ``Yes'', and should be relevant to ``main\_criteria'' and ``sub\_criteria''.\\
    - The ``main\_criteria'' should indicate the primary criteria of the ``question'', from the given collections.\\
    - The ``sub\_criteria'' should indicate more fine-grained criteria (represented as a concise noun phrase) belonging to ``main\_criteria''.\\
    - In each question of the ``checklist,'' it is NOT necessary to have exactly one question for each ``main\_criteria.'' That is, multiple questions can be generated for a single ``main\_criteria'' as needed.\\
    \\
    \lbrack Evaluation Checklists\rbrack\\
    Only generate the JSON-formatted evaluation checklists without any additional descriptions or explanations.\\
\end{prompt}

\begin{prompt}{Prompt Template used for \datasetName Evaluation: Quality Assessment}
    You will be provided with an image, a previous dialogue history between the user and the model, a reference answer that gets a score of 10, and an evaluation checklist. \\
    \\
    Your task is to evaluate the quality of the model's answer to the given dialogue history and image based on the provided evaluation checklist, which contains multiple questions. Compare the model's answer to the reference answer. \\
    \\
    \#\#\# Previous Dialogue History \\
    \textcolor{blue}{\texttt{\{dialogue\_history\}}} \\
    \\
    \#\#\# Model's Answer to evaluate: \\
    \textcolor{blue}{\texttt{\{model\_answer\}}} \\
    \\
    \#\#\# Reference Answer (Score 10): \\
    \textcolor{blue}{\texttt{\{reference\_answer\}}}\\
    \\
    \#\#\# Checklist (Evaluation Items)\\
    \textcolor{blue}{\texttt{\{checklist\}}}\\
    \\
    Please use this checklist to guide your evaluation, but do not limit your assessment to it. Compare the model's answer to the reference answer based on the checklist and the detailed criteria below. Scores should range from 1 to 10, where 1 indicates a very poor response and 10 signifies a perfect response. Here are more detailed criteria for the scores:\\
    \\
    \lbrack How well does the model perform overall in terms of accuracy, coherence, reasoning, informativeness, and user satisfaction?\rbrack\\
    - Score 1-2: The model's response is generally poor, with major issues in accuracy, coherence, reasoning, and relevance, making it unhelpful.\\
    - Score 3-4: The model's response is somewhat useful but contains noticeable flaws in accuracy, logical reasoning, or relevance that limit its effectiveness.\\
    - Score 5-6: The model's response is moderate, demonstrating reasonable accuracy, coherence, and informativeness, though some aspects may be improved.\\
    - Score 7-8: The model's response is strong, showing high accuracy, logical reasoning, and relevance with only minor weaknesses.\\
    - Score 9-10: The response is excellent, thoroughly accurate, and offers all necessary information to solve the user's problem.\\
    \\
    \#\#\# Output Format:\\
    Please provide your evaluation results in the following JSON format by filling in the placeholders in \[\]:\\
    json\\
    \{\\
        ``score'': ``[1~10]''\\
    \}\\
    \\
    \\
    Do not include any additional explanations or descriptions.\\
    \\
    Answer:
\end{prompt}

\begin{prompt}{Prompt Template used for \datasetName Evaluation: Checklist Completion Accuracy}
    You will be provided with an image, a previous dialogue history between the user and the model, a reference answer, and an evaluation checklist.\\
    \\
    Your task is to evaluate the quality of the model's answer to the given dialogue history and image based on the provided evaluation checklist, which contains multiple questions. Compare the model's answer to the reference answer.\\
    \\
    For each question, answer ``Yes'' or ``No.''\\
    \\
    \#\#\# Previous Dialogue History \\
    \textcolor{blue}{\texttt{\{dialogue\_history\}}} \\
    \\
    \#\#\# Model's Answer to evaluate: \\
    \textcolor{blue}{\texttt{\{model\_answer\}}} \\
    \\
    \#\#\# Reference Answer (Ground Truth): \\
    \textcolor{blue}{\texttt{\{reference\_answer\}}}\\
    \\
    \#\#\# Checklist (Evaluation Items)\\
    \textcolor{blue}{\texttt{\{checklist\}}}\\
    \\
    \#\#\# Output Format:\\
    Provide the final answer in the format of ``\textless Q \textgreater: \textless Yes or No \textgreater''. Do not include any additional explanations or descriptions.\\
    \\
    Answer:
\end{prompt}

\end{document}